%% file: main.tex
\documentclass{article}

\usepackage{arxiv}

\usepackage[utf8]{inputenc} 
\usepackage[T1]{fontenc}    
\usepackage{hyperref}       
\usepackage{url}            
\usepackage{booktabs}       
\usepackage{amsfonts}       
\usepackage{nicefrac}       
\usepackage{microtype}      
\usepackage{lipsum}
\usepackage{graphicx}
\graphicspath{ {./images/} }

\usepackage{booktabs}

\usepackage{amsmath,amsfonts}
\usepackage{array}
\usepackage{textcomp}
\usepackage{stfloats}
\usepackage{url}
\usepackage{verbatim}
\usepackage{graphicx}
\usepackage{cite}
\usepackage{hyperref}
\usepackage{url}
\usepackage{wrapfig}
\usepackage{graphicx}
\usepackage{booktabs}
\usepackage[switch]{lineno}
\usepackage{multirow}
\usepackage{cleveref}
\usepackage[ruled,vlined]{algorithm2e} 
\usepackage{float}
\usepackage{placeins}
\usepackage{adjustbox}

\newcommand{\code}[1]{\texttt{\footnotesize#1}}

\usepackage{tikz}
\usetikzlibrary{positioning,calc,fit,decorations.pathmorphing,shapes.geometric,shapes.gates.logic.US,calc}
\usepackage{tikz-3dplot}

\usepackage{pgfplots}
\usepackage{pgfplotstable}
\usepgfplotslibrary{groupplots}
\usepackage{wrapfig}
\usepackage{xcolor}
\usetikzlibrary{arrows.meta,positioning,fit,shadows}

\usepackage{threeparttable}
\usepackage{array}
\usepackage{graphicx}
\usepackage{wrapfig} 

\usepackage{algpseudocode}

\usepackage[utf8]{inputenc}
\usepackage[T1]{fontenc}
\usepackage{geometry}
\usepackage{times}
\usepackage{graphicx}
\usepackage{caption}
\usepackage{lipsum} 

\usepackage{tikz}
\usepackage{pgfplots}
\pgfplotsset{compat=1.18}
\usetikzlibrary{patterns} 

\usepackage{tcolorbox}
\usepackage{enumitem}
\tcbuselibrary{breakable}

\usepackage{xcolor}
\usepackage{mdframed}
\usepackage{listings}
\usepackage{subfigure} 
\usepackage{subcaption}

\definecolor{casegray}{gray}{0.96}

\newmdenv[
  backgroundcolor=casegray,
  linecolor=gray!35,
  linewidth=0.8pt,
  roundcorner=3pt,
  innertopmargin=0.6em,
  innerbottommargin=0.6em,
  innerleftmargin=0.8em,
  innerrightmargin=0.8em,
  skipabove=0.9em,
  skipbelow=0.9em
]{casebox}

\newmdenv[
  backgroundcolor=casegray,
  linecolor=gray!35,
  linewidth=0.3pt,
  roundcorner=2pt,
  innertopmargin=0.3em,
  innerbottommargin=0.3em,
  innerleftmargin=0.4em,
  innerrightmargin=0.4em,
  skipabove=0.5em,
  skipbelow=0.5em
]{compactcase}

\lstdefinestyle{logstyle}{
  basicstyle=\ttfamily\footnotesize,
  columns=fullflexible,
  breaklines=true,
  frame=none,
  showstringspaces=false
}

\usepackage{tikz}
\usetikzlibrary{arrows.meta, positioning}
\usepackage[dvipsnames,table,xcdraw]{xcolor}

\usepackage{xcolor}

\newcommand{\zhen}[1]{\textcolor{black}{#1}}
\newcommand{\zhenew}[1]{\textcolor{black}{#1}}
\usepackage{amsmath}
\usepackage{amssymb} 
\usepackage{dsfont}

\title{DLLM Agent: See Farther, Run Faster}

\author{\bfseries Hui-Ling Zhen$^{1,*}$, Weizhe Lin$^{1,*}$, Renxi Liu$^1$, Kai Han$^1$, Yiming Li$^1$, Yuchuan Tian$^{1,5}$, Hanting Chen$^1$, 
\\ \bfseries Xiaoguang Li$^1$, Xiaosong Li$^1$, Chen Chen$^1$, Xianzhi Yu$^1$, Mingxuan Yuan$^1$, Youliang Yan$^1$,  
\\ \bfseries Peifeng Qin$^1$, Jun Wang$^2$, Yu Wang$^3$, Dacheng Tao$^4$, Yunhe Wang$^1$ \\ 
\small \quad $^1$ Huawei Technologies Co., Ltd. \quad $^2$ UCL \quad $^3$ Tsinghua University \quad $^4$ NTU \quad $^5$ Peking University \\ $*$ Equal Contribution} 

\begin{document}
\maketitle
\begin{abstract}

\zhenew{Diffusion large language models (DLLMs) have emerged as an alternative to autoregressive (AR) decoding with appealing efficiency and modeling properties, yet their implications for agentic multi-step decision making remain underexplored. We ask a concrete question: when the generation paradigm is changed but the agent framework, supervision, and interaction budget are held fixed, do diffusion backbones induce systematically different planning and tool-use behaviors, and how do these differences interact with task structure?
We study this in a controlled setting by instantiating DLLM and AR backbones within the same agent workflow (DeepDiver-v2) and performing matched agent-oriented fine-tuning on identical trajectory data. Across BrowseComp-zh and open-ended tasks, we find that at comparable final accuracy, DLLM Agents are on average more than 30\% faster end-to-end, requiring fewer interaction rounds and tool invocations. Conditioned on success, DLLM trajectories exhibit substantially fewer conflict-correction events and reduced backtracking, indicating earlier global alignment of constraints.
Beyond efficiency, we uncover a structural interaction between generation paradigm and task topology. AR agents perform better on linear multi-hop chains that favor sequential dependency propagation, whereas diffusion-backed agents show clear advantages on constraint-intersection and evidence-reconciliation problems that require simultaneous multi-constraint coordination. This suggests that generation paradigms induce distinct structural inductive biases in multi-turn reasoning.
We further identify two practical considerations for deploying diffusion backbones in tool-using agents: increased susceptibility to structured tool-call violations and sensitivity to context–action masking alignment. Overall, our results indicate that diffusion backbones alter not only decoding speed but the allocation of reasoning commitments, leading to qualitatively different planning dynamics in agentic workflows.}

\end{abstract}

\input{arxiv/intro}
\input{arxiv/related}
\input{arxiv/method}
\input{arxiv/experiments}
\input{arxiv/analysis}

\input{arxiv/conclusion}



\bibliography{arxiv}
\bibliographystyle{IEEEtran}
\clearpage 

\appendix
\input{arxiv/appendix}

\end{document}

%% file: arxiv/intro.tex
\section{Introduction} \label{sec:intro}

Recent advances in large language models have led to the widespread adoption of language model agents, which extend single-shot generation to multi-turn reasoning, tool use, and structured workflows. Modern agent systems interact with external tools, maintain intermediate states, and revise decisions over multiple steps, forming complex action trajectories rather than isolated responses. As a result, agent performance is increasingly evaluated not only by final correctness, but also by efficiency, redundancy, and the structure of intermediate reasoning \cite{shi2025pangu,openpangu_deepdiver_v2}.

Despite this shift, most existing agent research implicitly assumes that agent behavior is primarily determined by training data and agent workflow design, while the underlying generation paradigm of the language model plays a secondary role \cite{xu2025comprehensive}. In practice, autoregressive (AR) decoding remains the de facto standard, and alternative generation paradigms are often treated as interchangeable components once embedded into a fixed agent framework. Under this assumption, improving agent performance is largely framed as a matter of better supervision, stronger base models, or more carefully engineered workflows \cite{ye2025dream,song2025seed}. However, this perspective overlooks a fundamental question: \emph{does the generation paradigm itself shape agent behavior in systematic ways that persist even under identical workflows and training data?}

In contrast to the AR paradigm, diffusion-based large language models (DLLMs) have emerged as a promising alternative paradigm for text generation.
Unlike AR models, which generate tokens sequentially, DLLMs formulate generation as an iterative denoising process over the entire sequence, enabling repeated revision and global coordination~\cite{wang2026top}. Prior work has demonstrated the viability of diffusion for language modeling through masked discrete diffusion and latent diffusion formulations, achieving competitive perplexity and generation quality while enabling parallel decoding and iterative refinement \cite{ye2025dream,sahoo2024simple}. Subsequent studies further emphasize diffusion models' ability to revise early decisions and support global semantic control, including latent planning and structured paragraph generation \cite{kang2025ladir,zhang2023planner}, soft-token evolution \cite{zhong2026beyond}, and latent reasoning through continuous--discrete diffusion \cite{zhou2025coevolutionary}.

Notably, prior research on diffusion language models has primarily focused on single-shot generation settings, leaving their role in agent-based reasoning workflows largely unexplored \cite{sahoo2024simple,ye2025dream,ma2024exploring}. This gap is important because agents are defined by trajectories of sequential decisions rather than isolated outputs. In practical agent workflows, both diffusion-based and autoregressive models generate actions step by step conditioned on accumulated interaction history \cite{geng2025webwatcher,li2025webthinker}. Nevertheless, their underlying training paradigms differ fundamentally. While AR models optimize strictly causal next-token prediction, DLLMs are trained through block-wise denoising with bidirectional attention within each block and iterative refinement across the sequence \cite{arriola2025block,bie2025llada2}.

Although bidirectional attention is locally constrained at the block level, the repeated denoising process enables \emph{implicit global coordination through repeated local refinement}, allowing the model to progressively align long-range structure across blocks \cite{wu2025fast,tian2026nexttokennextblockprincipledadaptation}. Within each block, bidirectional context further promotes more consistent and well-formed action representations \cite{tian2026nexttokennextblockprincipledadaptation}. Together, these properties suggest that DLLMs can implicitly capture long-horizon dependencies and downstream consequences when producing individual decisions, whereas AR generation emphasizes immediate local commitments \cite{sahoo2024simple,song2025seed}.
From this perspective, paradigm differences manifest not in the external workflow structure, but in the \emph{quality and global awareness of each local decision}. This can systematically influence planning efficiency, redundancy, and tool-use behavior, ultimately leading to distinct agent trajectories under otherwise identical workflows.

\begin{figure}
    \centering
    \includegraphics[width=13.6cm,height=8cm]
    {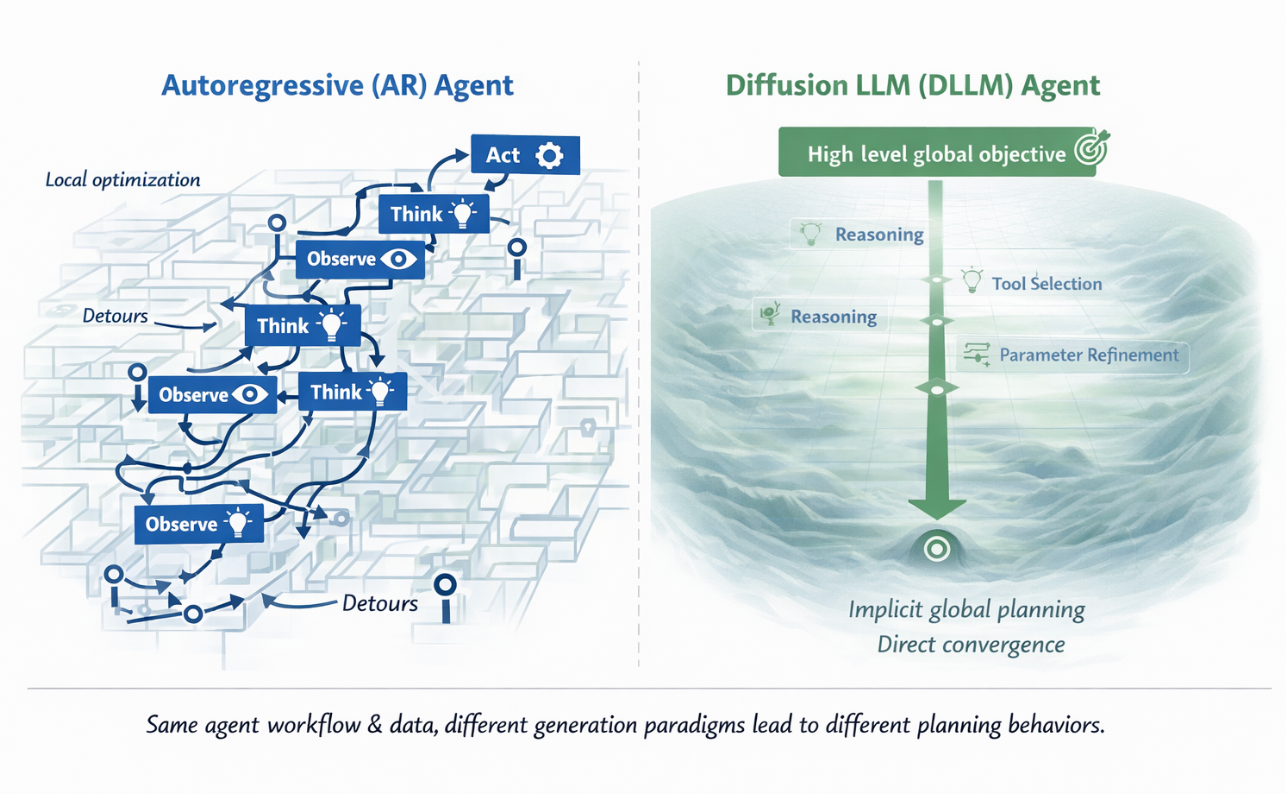}
    \caption{\textbf{Same agent workflow, different generation.} 
    Under identical tool-use workflows and training data, DLLM Agents (diffusion) tend to exhibit earlier trajectory-level convergence and less redundant interaction, while AR agents more often follow incremental, locally conditioned decisions that can lead to detours and repeated tool calls.}
    \label{fig:dllm_vs_ar_agent}
\end{figure}

In this paper, we investigate how different generation paradigms translate into agent behavior, and introduce \emph{Diffusion Large Language Model-based Agents (DLLM Agents)} by integrating diffusion models into standard agent workflows. We conduct a controlled comparison between diffusion-based and autoregressive language model agents by fine-tuning both paradigms on the same agent-oriented training data to equip them with comparable multi-turn dialogue and tool-use capabilities. 
Our analysis highlights consistent and systematic behavioral differences between DLLM Agents and their AR counterparts:

\begin{itemize}
    \item \textbf{Earlier convergence to effective trajectories.}
    On tasks with known solutions, DLLM Agents tend to identify viable action paths earlier, committing to correct plans with fewer exploratory detours.

    \item \textbf{Reduced redundancy.}
    AR agents more frequently exhibit repeated reasoning steps and unnecessary tool invocations, whereas diffusion agents produce more compact and decisive interaction traces.
    
    \item \textbf{Workflow-level efficiency.}
    The advantages of DLLM Agents extend beyond token-level parallelism, manifesting as fewer interaction steps and shorter overall agent trajectories.

    \item \textbf{Distinct exploration behaviors on open-ended tasks.}
    On questions without a single ground-truth answer, diffusion agents tend to exhibit broader exploration and higher-level revision patterns, while AR agents follow more incremental, locally conditioned decision processes.
\end{itemize}

Importantly, these differences persist under identical workflows and training data, indicating that they arise from the generation paradigm itself rather than superficial decoding speedups.

\paragraph{Contributions.}
In summary, this work makes the following contributions:
\begin{itemize}
    \item We present the first study of how diffusion and autoregressive generation paradigms differ at the level of agent behavior (Section~\ref{sec:method}).

    \item We demonstrate through extensive case studies (Section~\ref{sec:exp}) and open-ended tasks that DLLM Agents achieve superior planning efficiency and reduced redundancy compared to AR-based agents.

    \item We analyze attention dynamics from an agent workflow perspective (Section~\ref{sec:analysis}), revealing that DLLM agents leverage early-stage global attention for action coordination, while AR agents rely on stepwise, token-level decision making.
\end{itemize}
Together, our findings highlight diffusion-based generation as a promising paradigm for building more efficient and structured language model agents. Conclusions and future works are in Section~\ref{sec:conclusion}. 

%% file: arxiv/related.tex
\section{Related Work}\label{sec:related}

\paragraph{LLM Agents \& Tool-use Workflows.}
Language-model agents extend single-shot generation into multi-step interaction loops that combine reasoning, state tracking, and external tool use \cite{zheng2025deepresearcher,xu2025comprehensive}. Recent systems emphasize structured workflows (e.g., plan-act-reflect), explicit tool APIs, and environment feedback, shifting evaluation from final answer quality to trajectory-level properties such as the number of turns, tool-call cost, and recovery behaviors in long-horizon tasks \cite{openpangu_deepdiver_v2,shi2025pangu,li2025webthinker}. In parallel, recent benchmarks and agentic evaluations have begun to operationalize tool-use competence and long-horizon reliability in more realistic settings (e.g., software engineering, web navigation, and general tool ecosystems), further highlighting the importance of workflow design and interaction efficiency \cite{lin2025towards,jiang2024webarena}. Beyond prompt-level scaffolding, these workflows increasingly resemble general decision-making pipelines where the model must coordinate intermediate states, choose actions, and revise plans under partial observability \cite{huang2025deep,jiang2024webarena}. Our work follows this line by using a fixed agent framework (\emph{i.e.}, DeepDiver~\cite{wan2025deepresearch}) and DeppDiver-v2\cite{openpangu_deepdiver_v2} to isolate how the \emph{generation paradigm} affects agent trajectories under controlled workflows and training data. 

\paragraph{Decoding/Search for AR Agents.}
A complementary thread improves agent behavior by augmenting autoregressive (AR) generation with inference-time search and revision. These methods typically keep the AR backbone unchanged while modifying how candidates are proposed, evaluated, and aggregated. For example, sampling multiple reasoning traces and selecting consistent solutions (self-consistency), using explicit verification or reflection to iteratively refine outputs, and applying tree-structured search with learned or heuristic evaluators to explore alternative action sequences \cite{ye2025agentfold,pei2025scope}. Recent work also revisits RL-style test-time control and search for stronger decision-making, including planning-oriented rollouts and reward-guided inference that can be applied to multi-step problem solving and agent settings \cite{shinn2024reflexion,lin2025trimr,tang2026rethinker}. While such approaches demonstrate that inference procedures can substantially influence agent reliability without changing the underlying modeling paradigm, our focus is orthogonal: we hold the agent workflow constant and compare diffusion vs.\ AR backbones under matched agent-oriented fine-tuning, asking whether the \emph{generation paradigm itself} induces systematic differences in planning, redundancy, and trajectory efficiency \cite{lin2025towards}.

\paragraph{Diffusion LLMs for Text Generation.}
Diffusion-based language modeling has re-emerged as a competitive alternative to AR generation by formulating text generation as an iterative denoising process over an entire sequence, enabling global coordination and repeated revision. Recent work shows that discrete or latent diffusion formulations can reach strong generation quality while offering parallel decoding and iterative refinement \cite{sahoo2024simple,ye2025dream,tian2026nexttokennextblockprincipledadaptation}. Building on these foundations, subsequent studies explore diffusion's ability to support higher-level semantic control and structured generation, such as latent planning mechanisms and paragraph-level organization \cite{kang2025ladir,ma2026diffusiondiffusionreclaimingglobal}, as well as continuous-discrete formulations that facilitate latent reasoning and progressive refinement \cite{zhou2025coevolutionary}. However, this literature has largely remained within single-shot (or single-episode) generation settings and has not been systematically evaluated in tool-using, multi-turn agent workflows where behavior is defined by action trajectories rather than one-off outputs.

\paragraph{Agent Behavior Analysis \& Attention/Trajectory Metrics.}
As agents become more widely deployed, an increasing number of studies analyze agent behavior beyond final accuracy, focusing on process metrics such as redundancy, tool-use efficiency, trajectory length, and failure modes across open-ended and interactive tasks \cite{wu2025webdancer,geng2025webwatcher,zhang2025evolvesearch}. Recent benchmark-driven analyses in realistic environments (e.g., web navigation and software engineering) also surface trajectory-level bottlenecks such as unnecessary actions, brittle recovery, and high interaction cost, motivating metrics that capture efficiency and robustness rather than only endpoint success \cite{jiang2024webarena,luo2025graph}. Another line of work studies internal model dynamics (e.g., attention patterns and information flow) to explain how models coordinate decisions over multi-step interactions, though such analyses are often token-centric or single-generation-centric rather than aligned to explicit workflow stages \cite{shi2025pangu,shinn2024reflexion,li2025deepagent}. Against the backdrop of growing interest in DLLMs as autoregressive backbone alternatives for low-latency real-time agentic interaction, recent work conducts a comprehensive evaluation of representative DLLMs on embodied and tool-calling agent paradigms, and uncovers that current DLLMs cannot act as reliable agentic backbones due to systematic failures in long-horizon planning and symbolic precision maintenance under diffusion noise \cite{lu2026bitter}. Our analysis complements these efforts by aligning attention dynamics to the agent workflow and comparing diffusion vs.\ AR agents under controlled conditions, aiming to reveal paradigm-specific coordination mechanisms that manifest at the trajectory level. 

%% file: arxiv/method.tex
\section{Method}
\label{sec:method}

\subsection{Overview: Controlled Agent Comparison}
\label{sec:method_overview}

Our goal is to isolate how the \emph{generation paradigm} of the policy backbone—autoregressive (AR) versus diffusion language models (DLLMs)—affects tool-using agent behavior in complex multi-turn workflows.
To this end, we conduct a controlled comparison in which the entire agent pipeline is kept fixed and only the backbone used to generate each action is swapped.

Across both agents, we match: (i) the multi-agent orchestration framework; (ii) the tool set and APIs; (iii) prompt templates and state serialization; (iv) training data and optimization budgets; and (v) evaluation protocols and interaction budgets.
Under these controls, the sole varying factor is the intra-step generation mechanism: AR decoding produces actions through left-to-right token prediction with irreversible commitment to early tokens, whereas DLLMs produce actions through iterative denoising and global refinement.

This controlled setup allows us to attribute differences in convergence to correct action trajectories, redundancy in intermediate reasoning and tool calls, and overall workflow efficiency directly to the backbone generation paradigm.


\subsection{Agent Workflow and Tool Interface}
\label{sec:workflow_interface}
In this work, we employ multi-agent deep research (\emph{i.e.}, DeepDiver-v2~\cite{openpangu_deepdiver_v2}) as a representative scenario and establish it as our standardized agent workflow for experimental validation.
DeepDiver-v2 operates through hierarchical orchestration, where a \emph{Planner Agent} decomposes the global objective into iterative sub-goals and dynamically coordinates specialized agents.
Information Seeker agents are responsible for interacting with external knowledge sources and exploration tools, while Writer agents synthesize retrieved evidence and produce structured outputs.

The system is equipped with a heterogeneous tool ecosystem, encompassing both cognitive operators and environment-facing interfaces.
Cognitive tools include operators such as \texttt{think} and \texttt{reflect} for explicit reasoning and self-correction, while environment-facing tools include \texttt{batch\_web\_search}, \texttt{url\_crawler}, \texttt{document\_qa}, \texttt{file\_read}, and \texttt{file\_write}, enabling information retrieval, document analysis, and external state manipulation.

A typical agent iteration follows a structured \emph{think--act--observe} loop:
the Planner Agent first reasons over the current workflow context to formulate the next sub-goal,
invokes the appropriate tool through a specialized agent,
and then integrates the returned observation into the shared history before proceeding to the next decision.

We abstract each coordinated multi-agent decision as a structured action:
\begin{equation}
a_t \in \{\mathrm{ToolCall}(u_t,\theta_t),\ \mathrm{Terminate}(y)\},
\end{equation}
where $u_t$ denotes the selected tool and $\theta_t$ its parameters. Concrete operations such as web search, document QA, and file I/O correspond to particular ToolCall instances. $\mathrm{Terminate}(y)$ outputs the final response and ends the episode.

Executing an action yields an observation $o_t$, which is appended to the workflow history and conditions subsequent decisions.


\subsection{Trajectory and Backbone Policy Formulation}
\label{sec:traj_formulation}

\paragraph{Workflow state and trajectory.}
Given a user query $q$, the global execution context at round $t$ is summarized by the accumulated history:
\begin{equation}
\mathcal{H}_{t-1} = \{ q, (x_1,o_1), \dots, (x_{t-1},o_{t-1}) \},
\end{equation}
where $x_i$ is the textual serialization of the structured action issued at round $i$ (including reasoning fields and tool invocations), and $o_i$ is the corresponding observation.

An episode produces a workflow trajectory:
\begin{equation}
\tau = \bigl\{(\mathcal{H}_{t-1}, x_t, a_t, o_t)\bigr\}_{t=1}^{T^\star},
\end{equation}
with termination step $T^\star$.

This trajectory-level formulation allows us to analyze agent behavior at the workflow level, capturing not only final task success but also intermediate efficiency indicators such as the number of tool calls, redundant reasoning steps, and convergence speed to viable action paths.

\paragraph{AR policy factorization.}
Under an autoregressive backbone, each round is generated through a left-to-right factorized process:
\begin{align}
r_t &\sim \pi_{\mathrm{AR}}(r \mid \mathcal{H}_{t-1}), \ \ \ \ a_t \sim \pi_{\mathrm{AR}}(a \mid \mathcal{H}_{t-1}, r_t),
\end{align}
where $r_t$ denotes intermediate reasoning tokens and $a_t$ the structured action outcome.
This sequential commitment can amplify verbosity and often necessitates corrective actions across multiple rounds.

\paragraph{Diffusion-based policy refinement.}
With a diffusion backbone, the coordinated decision for round $t$ is produced through iterative refinement over the entire structured action segment.
Let $x_t$ denote the serialized action proposal.
The DLLM defines a denoising process:
\begin{equation}
x_t^{(0)} \rightarrow x_t^{(1)} \rightarrow \cdots \rightarrow x_t^{(K_t)},
\quad \text{where } K_t \text{ is adaptively determined by confidence-gated decoding.}
\end{equation}

\noindent\emph{Implementation note:} while this formulation treats $x_t$ as a single sequence for clarity, the denoising process is applied over contiguous token blocks in practice, enabling scalable refinement under the same conditional objective.

The final refined output $x_t^{(K)}$ is parsed into the structured action $a_t$ and executed.
Across backbones, the orchestration logic and tool semantics remain unchanged; only the generation mechanism differs.


\subsection{Agent-oriented Fine-tuning for DLLM}
\label{sec:agent_ft}

We adapt DLLMs as policy backbones through agent-oriented fine-tuning that directly targets structured action decisions.
Although the supervision source---agent trajectories---matches standard agent SFT used for AR models, the learning objective differs fundamentally.

\paragraph{Action-level supervision.}
We construct a dataset of rollout pairs:
\begin{equation}
\tau = \bigl\{(\mathcal{H}_{t-1}, x_t)\bigr\}_{t=1}^{T^\star},
\end{equation}
\noindent where $\mathcal{H}_{t-1}$ denotes the interaction history up to round $t{-}1$ (including user messages, intermediate tool observations, and prior agent actions), $x_t$ is the agent's \emph{action segment} at round $t$, and $T^\star$ is the number of decision rounds in the trajectory.
Each action segment $x_t$ contains (i) optional reasoning tokens and (ii) exactly one structured outcome: either a \texttt{ToolCall} with arguments or a terminal response.
Using action-level targets ensures that the model learns to produce coherent, complete decisions aligned with the agent workflow, rather than generic dialogue continuations.

\paragraph{Conditional denoising objective.}
Let $x \triangleq x_t$ and $c \triangleq \mathcal{H}_{t-1}$.
DLLMs corrupt the target action segment with a \emph{noise-level-indexed} corruption operator $\mathcal{C}_k$:
\begin{equation}
\tilde{x} = \mathcal{C}_k(x),
\end{equation}
\noindent where $k \in \{1,\dots,K\}$ indexes the corruption (noise) level, $K$ is the maximum number of diffusion steps (or discrete noise levels) used in training, and $\mathcal{C}_k(\cdot)$ is a step-dependent stochastic operator that increasingly perturbs $x$ as $k$ grows (e.g., by masking or replacing a larger fraction of tokens or token spans at higher $k$).

The primary diffusion objective trains the model to recover the clean action segment conditioned on the corrupted input, the context, and the noise level:
\begin{equation}
\mathcal{L}_{\mathrm{MDM}}
= \mathbb{E}_{(c,x)\sim \tau,\; k \sim \mathcal{U}(\{1,\dots,K\})}
\Bigl[- \log p_{\theta}(x \mid \tilde{x}, c, k)\Bigr],
\label{eq:dllm_denoise_loss}
\end{equation}
\noindent where $\tilde{x}=\mathcal{C}_k(x)$, $p_{\theta}$ denotes the parameterized conditional distribution of the clean action segment, and $\mathcal{U}$ is the uniform distribution over discrete noise levels.\footnote{Other sampling schedules for $k$ are possible; we adopt uniform sampling for simplicity.}

In addition to the denoising objective, we incorporate a standard autoregressive cross-entropy loss over the clean action segment to stabilize training and preserve strong left-to-right generation capability:
\begin{equation}
\mathcal{L}_{\mathrm{AR}}
= \mathbb{E}_{(c,x)\sim \tau}
\Bigl[
-\!\!\sum_{i=1}^{|x|}
\log p_{\theta}(x_i \mid x_{<i}, c)
\Bigr].
\end{equation}

The overall training objective is formulated as a weighted combination:
\begin{equation}
\mathcal{L}
= \mathcal{L}_{\mathrm{MDM}} + \lambda \, \mathcal{L}_{\mathrm{AR}},
\end{equation}
\zhenew{where $\lambda = 0.5$ balances the diffusion modeling loss and the autoregressive (AR) guidance term. 
This choice follows prior AR-to-DLM adaptation studies~\cite{tian2026nexttokennextblockprincipledadaptation}, 
which show that maintaining a comparable AR objective during transition stabilizes optimization and preserves long-context reasoning capabilities inherited from pretrained AR models. 
In particular, equal weighting provides a symmetric regularization effect: 
the diffusion objective encourages global refinement and parallel denoising over token blocks, 
while the AR term anchors training to the original next-token prediction manifold, preventing degeneration during early-stage adaptation.
In practice, both corruption and denoising are applied over contiguous token blocks (spans) under a block-diffusion formulation, 
while preserving the same conditional objective over the prefix context. 
Compared with naive continuation-based diffusion training on generic dialogue data, 
this agent-oriented formulation better aligns learning with the discrete decision structure of tool-using workflows. 
Such alignment is especially critical under strict interaction budgets, 
where stable planning trajectories and reduced branching are necessary for effective execution. }

\zhen{Since our diffusion agent uses a block-structured formulation, the denoising problem is not fully “global” over the entire segment: within each block there remains an intrinsic left-to-right dependency (i.e., later tokens in the block are naturally conditioned on earlier tokens). In this setting, optimizing only the conditional denoising objective can weaken the model’s token-level causal consistency and lead to less stable training, since the model is never explicitly required to be a strong next-token predictor on clean sequences. We therefore add an auxiliary autoregressive (AR) cross-entropy term on the uncorrupted action segment. This AR loss serves as a regularizer that preserves the model’s left-to-right generation capability and encourages coherent local sequencing inside each block, while the diffusion-style denoising loss provides robustness to corruption and enables multi-step refinement across discrete noise levels. The combined objective thus matches the hybrid nature of block diffusion—partially denoising-based, yet still respecting causal structure—yielding improved stability and better sequential fidelity in action generation.}

\zhen{Although both AR and DLLM agents are trained on similar trajectory data, their generation mechanisms differ fundamentally. 
AR models factorize the joint probability of an action sequence into a left-to-right product of conditional next-token distributions, optimizing a token-level likelihood objective. This induces a strictly sequential generation process in which each token is committed once generated.  
In contrast, DLLMs optimize a denoising objective over partially corrupted spans, learning to iteratively refine entire segments of text. 
Generation proceeds through multiple refinement steps, where tokens can be globally adjusted before convergence. 
This bidirectional conditioning and iterative correction mechanism introduces a different inductive bias: instead of committing to early token decisions, the model can revise them during refinement.}


\subsection{Diffusion Masking for Multi-turn Agent Contexts}
\label{sec:diffusion_masking}

In multi-turn agent workflows, inference always conditions on a fully clean history and applies diffusion only to the next-step action segment.
However, standard DLLM training typically corrupts entire sequences and allows bidirectional interactions across fixed token blocks, introducing severe train--inference mismatches. Such mismatches are particularly problematic in multi-turn settings, where the context encodes strict causal dependencies between previous agent outputs and subsequent tool responses, whereas these dependencies are absent in single-turn generation.

We introduce two complementary masking designs to address the mismatch problem in multi-turn settings. To explicitly align training with agent inference behavior, we decompose each input into a clean context prefix and a target action span: $c_t \;\Vert\; x_t$, where $c_t=\mathcal{H}_{t-1}$ encodes the full interaction history and $x_t$ corresponds to the next structured action.

\paragraph{Context-clean corruption.}

Let $\mathcal{I}_c$ and $\mathcal{I}_x$ denote the token index sets corresponding to the clean context prefix $c_t$ and the target action span $x_t$, respectively, with $\mathcal{I}_c \cap \mathcal{I}_x = \emptyset$. 
We define a step-dependent corruption operator $\mathcal{C}_k(\cdot)$ that applies masking or noise to a subset of tokens according to the diffusion step $k$.
Under context-clean corruption, diffusion noise is applied exclusively to the action span:
\begin{equation}
\tilde{x}_{t,i} =
\begin{cases}
\mathcal{C}_k(x_{t,i}), & i \in \mathcal{I}_x, \\
x_{t,i}, & i \in \mathcal{I}_c,
\end{cases}
\end{equation}
yielding the denoising objective
\begin{equation}
\mathcal{L}_{\mathrm{MDM}}
=
\mathbb{E}_{(c_t,x_t)\sim \tau,\; k \sim \mathcal{U}(\{1,\dots,K\})}
\Bigl[
-\!\!\sum_{i \in \mathcal{I}_x}
\log p_{\theta}(x_{t,i} \mid \tilde{x}_t, c_t, k)
\Bigr].
\end{equation}

This design ensures that the context prefix always remains fully observed during training, matching the inference-time input distribution in multi-turn workflows.

\paragraph{Span-aware attention alignment.}

While context-clean corruption aligns the input distribution, standard block diffusion still applies bidirectional self-attention uniformly within fixed-length blocks.
When a block overlaps the boundary between context and action tokens, this introduces attention paths that do not exist at inference.

To remove such spurious interactions, we introduce a span-aware attention constraint.

Let $\mathcal{I}^{\mathrm{loss}}=\mathcal{I}_x$ denote token positions participating in the denoising objective (i.e., action-span tokens), and let $\mathcal{I}^{\mathrm{ctx}}=\mathcal{I}_c$ denote non-trainable context positions.

For any attention query position $i \in \mathcal{I}^{\mathrm{loss}}$, we restrict attention to:

\begin{itemize}
\item all earlier context tokens $j \in \mathcal{I}^{\mathrm{ctx}}$ with $j < i$ (causal access), and
\item all tokens within the action span $\mathcal{I}^{\mathrm{loss}}$ (bidirectional diffusion within the span).
\end{itemize}

Formally, we define the attention mask $M(i,j)$ as:
\begin{equation}
M(i,j)=
\begin{cases}
0, & i \in \mathcal{I}^{\mathrm{loss}},\ j \in \mathcal{I}^{\mathrm{ctx}},\ j \ge i, \\
1, & \text{otherwise},
\end{cases}
\end{equation}
where $M(i,j)=0$ removes the attention edge from query $i$ to key $j$.

This constraint preserves valid causal context conditioning while eliminating non-existent bidirectional attention paths introduced by blockwise diffusion.

\paragraph{Orthogonality.}
Context-clean corruption aligns the input distribution, while span-aware attention aligns the attention structure.
Both can be enabled independently or jointly.


\subsection{Running a DLLM Agent in the Workflow}
\label{sec:dllm_inference}

\paragraph{Inference-time refinement.}
At round $t$, the history $\mathcal{H}_{t-1}$ is serialized into conditioning context $c_t$.
Starting from an initialized action canvas $x_t^{(0)}$, the DLLM applies the refinement process in Section~\ref{sec:traj_formulation} for $K$ steps to obtain $x_t=x_t^{(K)}$.

This procedure jointly refines reasoning structure, tool selection, argument construction, and termination decisions before execution.

\paragraph{Per-round adaptive decoding (DLLM only).}
The DLLM generates each action segment using a standard confidence-gated decoding procedure.
At each iteration, masked tokens whose prediction confidence exceeds a threshold $\tau$ (we use $\tau=0.9$) are committed and the process repeats until all tokens within the decoding window are resolved.
This yields an adaptive number of refinement iterations per round while avoiding unnecessary computation once the action becomes confident.

\paragraph{Action execution and recovery.}
After denoising, the refined action segment $x_t$ is parsed into a discrete executable action:
\begin{equation}
a_t \in \{\mathrm{ToolCall}(u_t,\theta_t),\ \mathrm{Terminate}(y_t)\},
\end{equation}
\noindent where $u_t$ is the tool identifier, $\theta_t$ is the structured argument payload, and $y_t$ is the final natural-language answer when the episode terminates. Malformed outputs are handled using identical recovery strategies across backbones to avoid confounding effects.  
\zhenew{ We emphasize that no explicit unified efficiency objective (e.g., a weighted combination of tool calls, tracking length, or interaction turns) is introduced during training. 
All efficiency-related quantities reported in Table~\ref{tab:browsecomp_plan} are purely evaluation metrics computed at the episode level. 
The training objective remains strictly likelihood-based (Eq.~\ref{eq:dllm_denoise_loss} and the combined loss thereafter), without reward shaping or auxiliary penalties on tool usage or trajectory length. Therefore, differences in tool-call frequency, interaction turns, or planning stability should be understood as emergent properties of the underlying generation mechanisms under identical supervision and interaction budgets, rather than as artifacts of explicitly optimized efficiency terms.}


\subsection{Evaluation Protocol and Budget Constraints}
\label{sec:eval_protocol}

Both AR and DLLM agents operate under identical interaction budgets, including a strict global context-length cap (up to 32K tokens), maximum number of rounds $T_{\max}$, maximum number of tool calls $C_{\max}$, and total generated token limits for agent-produced text. 
If an episode would exceed these constraints, termination follows the same fallback policy for both backbones.

%% file: arxiv/experiments.tex
\section{Experiments}  \label{sec:exp}

\subsection{Experimental Setup}
\label{sec:exp_setup}

This section describes the evaluation protocol for comparing AR and diffusion-based agents within the same DeepDiver workflow, with careful control over interaction budgets and context-length constraints.

\paragraph{Tasks and datasets.}
We evaluate agent performance in two complementary settings.

\textit{BrowseComp-zh.}
We adopt the public \textsc{BrowseComp-zh} benchmark~\cite{zhou2025browsecomp}, which requires multi-turn web browsing, iterative retrieval, and reasoning over dynamically collected evidence to answer user queries.
The benchmark naturally induces long interaction histories due to repeated tool use and document accumulation.

\textit{Open-ended qualitative prompts.}
In addition to benchmark evaluation, we include a small suite of open-ended, multi-step tasks (e.g., document drafting and revision) that are intended to elicit long-horizon planning, tool chaining, and stable action execution. Unlike standard benchmark items, these prompts often admit multiple acceptable solutions and are therefore not scored with a single “correct/incorrect” label. Instead, we use them purely for qualitative analysis, inspecting full agent trajectories to compare how different agents work: their planning and decomposition strategies, the structure and frequency of tool calls, iteration and self-correction behavior, and characteristic failure modes (e.g., drift, premature termination, redundancy, or instability). This setting complements automatic metrics by revealing behavioral differences and process-level patterns that are difficult to capture with aggregate scores.

\paragraph{Compared agent variants.}
All agents are instantiated within the same DeepDiver iterative RAG workflow, sharing identical orchestration logic, tool APIs, prompt templates, and action parsers.

We compare:
\begin{itemize}
    \item \textbf{AR Agent.} A DeepDiver agent with an AR LLM (openPangu-Embedded-7B \cite{openpangu-embedded-7b}) backbone that generates each action span through left-to-right decoding. The AR backbone is fine-tuned on the same trajectory dataset under matched optimization budgets.

    \item \textbf{DLLM Agent.} A DeepDiver agent with a diffusion language model (openPangu-R-7B-Diffusion \cite{tian2026nexttokennextblockprincipledadaptation}) backbone that generates each action span via iterative denoising and refinement. The DLLM is trained using the agent-oriented fine-tuning objective and multi-turn masking strategies described in Section~\ref{sec:agent_ft} and Section~\ref{sec:diffusion_masking}.
\end{itemize}

\paragraph{Shared interaction budgets and constraints.}
To ensure a controlled comparison, we impose identical evaluation budgets across all agents:

\begin{itemize}
    \item \textbf{Context-length cap.} All agents operate under a maximum context window of 32K tokens. If adding the next interaction would exceed this limit, the episode is terminated using the same fallback policy across methods (e.g., returning the most recent valid $\mathrm{Terminate}(y)$ action).

    \item \textbf{Interaction cap.} The maximum number of interaction rounds is fixed to $T_{\max}=15$ for all agents.

    \item \textbf{Tool-use cap.} We apply the same upper bound on the number of ToolCall invocations per episode. Once the budget is exhausted, the agent is prompted to produce a terminal action.

\end{itemize}

Both models were trained for 5 epochs with a decaying learning rate from $5e^{-6}$ to $0$.
All action segments are extracted using the same \texttt{BEGIN\_ACTION}/\texttt{END\_ACTION} delimiters and parsed with identical deterministic rules to avoid confounding effects from formatting differences.


\subsection{Benchmark Evaluation and Evaluation Protocol}
\label{sec:benchmark_eval}

We benchmark AR and DLLM agent on \textsc{BrowseComp-zh} under the shared interaction budgets described above, ensuring identical training data, context limits, and tool access. 
Our evaluation is designed to compare not only end-task accuracy but also trajectory-level properties that affect practical agent deployment.

\paragraph{End-task performance.}
We report the standard benchmark success metric, which measures the correctness of the final answer after completing multi-turn browsing and reasoning. 
This metric captures the overall problem-solving capability of the agent under fixed interaction constraints.

Beyond final accuracy, we analyze trajectory-level statistics that reflect how efficiently and robustly different backbones operate in multi-turn settings. 
Specifically, we measure:

\begin{itemize}
    \item Interaction rounds: the number of turns required to reach a terminal decision;
    \item Tool usage: the number of \texttt{ToolCall} invocations per episode;
    \item Redundancy: repeated tool calls or unnecessary intermediate reasoning steps;
    \item Formatting robustness: the invalid action rate, defined as the fraction of episodes containing at least one unparsable action span (e.g., missing delimiters, malformed tool calls, or schema violations).
\end{itemize}

Together, these metrics characterize how stably and economically different generation paradigms navigate complex, long-horizon browsing workflows.

\paragraph{Qualitative trajectory analysis.}
For open-ended prompts, we additionally present representative agent trajectories to illustrate differences in planning structure, tool selection patterns, and long-horizon coherence between AR and diffusion-based agents. 
These case studies complement quantitative metrics by exposing backbone-specific behavioral tendencies. 
Unless otherwise specified, all quantitative results are reported on a 110-question subset uniformly sampled from BrowseComp-zh~\cite{zhou2025browsecomp}. \zhenew{We also analyze calibration error~\cite{zhou2025browsecomp} at the \emph{episode level}. 
For each completed episode, the agent produces a scalar confidence score corresponding to its predicted probability that the final answer is correct. 
Formally, let $p_j \in [0,1]$ denote the predicted probability of final-answer correctness for episode $j$, and let $y_j \in \{0,1\}$ indicate whether the final answer is correct. To compute the expected calibration error (ECE), we partition the predicted probabilities into five bins: $[0,0.2)$, $[0.2,0.4)$, $[0.4,0.6)$, $[0.6,0.8)$, and $[0.8,1.0]$. 
For each bin $i$, we compute the empirical accuracy $acc(i) = \frac{1}{n_i} \sum_{j \in \mathcal{B}_i} y_j$, and the average predicted confidence $conf(i) = \frac{1}{n_i} \sum_{j \in \mathcal{B}_i} p_j$, where $\mathcal{B}_i$ is the set of episodes whose predicted confidence falls into bin $i$, and $n_i = |\mathcal{B}_i|$. 
The ECE is then defined as
\[
\mathrm{ECE}=\sum_{i=1}^{B}\frac{n_i}{N}\left|acc(i)-conf(i)\right|,
\]
where $B$ is the number of bins and $N$ is the total number of evaluated episodes. 
This definition corresponds to the 5-bin episode-level calibration metric reported in Table~\ref{tab:browsecomp_plan}. }

\begin{figure}[htbp]
    \centering
    \includegraphics[width=0.8\linewidth]{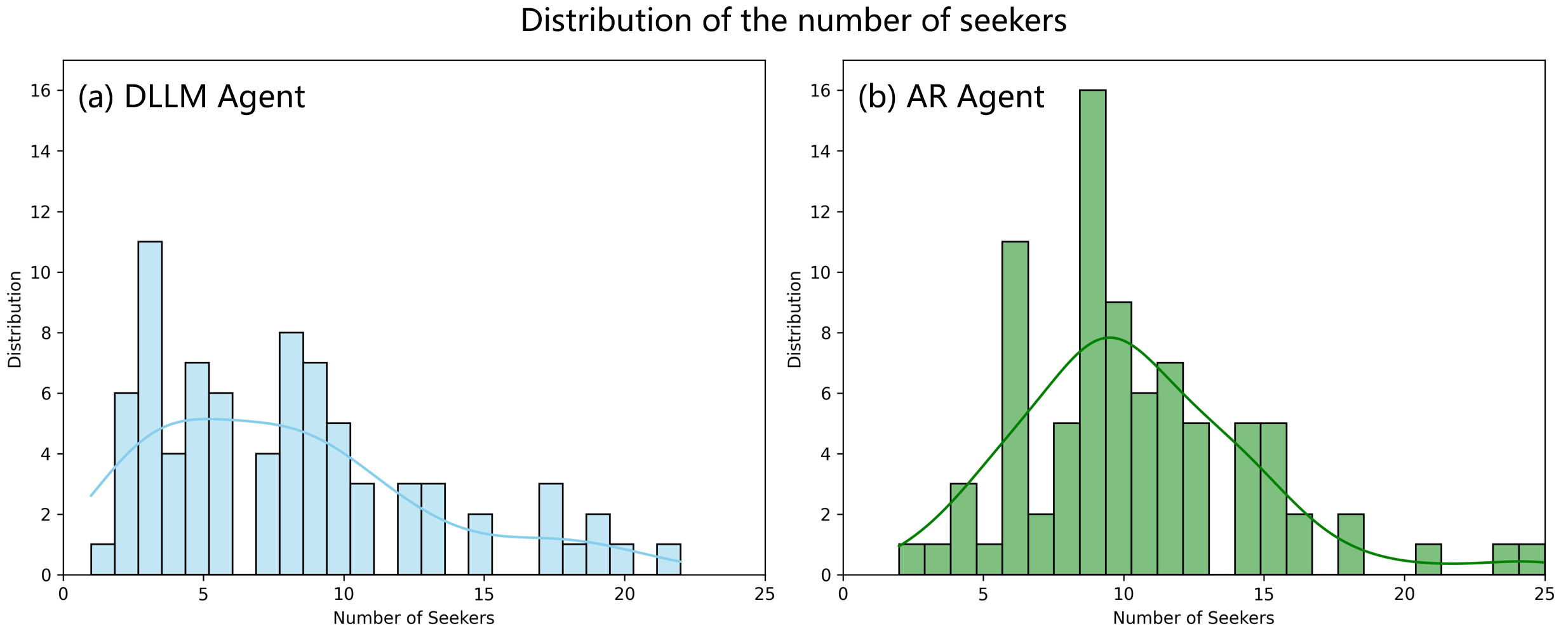}
    \caption{(a) and (b) show the distributions of the number of times the information seeker is invoked by the DLLM agent and the AR agent, respectively. The DLLM agent invokes the seeker significantly fewer times than the AR Agent.
}
    \label{fig:nseeker}
\end{figure}

\begin{table}[htbp]
    \centering
    \small
    \setlength{\tabcolsep}{4.2pt}
    \resizebox{\linewidth}{!}{
    \begin{tabular}{lccccccc}
        \toprule
        \textbf{Method} &
        \textbf{Accuracy} $\uparrow$ &
        \textbf{Calibration Error} $\downarrow$  &
        \textbf{Tool Calls} $\downarrow$ &
        \textbf{Turns Used} $\downarrow$ &
        \textbf{Invalid Action Rate} $\downarrow$ &
        \textbf{Plan@Ep} $\uparrow$ &
        \textbf{PlanRate} $\uparrow$ \\
        \midrule
        AR Agent & 15.5\% & 59\% & 7.5 & 14.8 & 1.9\% & 0.55 & 38\%  \\
        DLLM Agent & 15.5\% & 66\% & 6.7 & 13.0 & 6.4\% & 0.92 & 62\%   \\
        ~~\textit{w/o context-clean corruption}  & 14.5\% & 68\% & 6.8 & 15.0 & 6.2\% & 0.85 & 57\% \\
        ~~\textit{w/o span-aware attention alignment} & 14.5\% & 67\% & 7.1 & 14.6 & 7.1\% & 0.79 & 53\%  \\
        \bottomrule
    \end{tabular}}
    \vspace{0.2cm}
    \caption{ Comparison of AR and diffusion-based (DLLM) agents on the BrowseComp-zh benchmark under the same 32K context limit and the same maximum interaction budget; ablations remove individual masking components from DLLM. Metrics are averaged over episodes. \textbf{Accuracy} is final-answer correctness; \textbf{Expected Calibation Error (ECE)} is the 5-bin expected calibration error of episode-level confidence for final correctness (lower is better). \textbf{Invalid Action Rate} is the fraction of episodes containing at least one unparsable tool call under a strict schema parser. \textbf{Plan@Ep} is the mean number of detected global-planning spans per episode, and \textbf{PlanRate} is the fraction of episodes with at least one detected planning span. 
}
    \label{tab:browsecomp_plan}
\end{table}

\paragraph{Overall performance, efficiency, and structural planning behavior.} 
Table~\ref{tab:browsecomp_plan} summarizes benchmark accuracy together with efficiency, stability, and planning-related metrics under the same 32K context constraint and a shared maximum interaction budget.
Under matched decoding budgets, \textbf{the DLLM agent achieves the same final-answer accuracy as the AR agent (15.5\%), while requiring fewer intermediate interactions and reducing end-to-end inference time by over 30\%. }
Specifically, DLLM makes fewer tool calls (6.7 vs.\ 7.5) and uses fewer turns (13.0 vs.\ 14.8), resulting in lower end-to-end latency (Figure~\ref{fig:nseeker}). Beyond raw interaction counts, we examine \emph{how} the two agents allocate their budgets.
DLLM exhibits substantially higher global-planning coverage and intensity (PlanRate: 62\% vs.\ 38\%; Plan@Ep: 0.92 vs.\ 0.55), indicating more frequent explicit episode-level planning behaviors such as subgoal decomposition and tool allocation before execution.
This difference is consistent with diffusion-style generation, which refines entire spans holistically and encourages coordinated action planning, whereas AR decoding proceeds left-to-right with more incremental commitments.
Thus, under matched success rates, DLLM achieves comparable task performance with more compact yet more plan-driven trajectories.

However, this compactness introduces a structural trade-off.
DLLM shows a higher invalid action rate (6.4\% vs.\ 1.9\%), suggesting reduced local structured-action reliability despite improved global coordination.
The ablations in Table~\ref{tab:browsecomp_plan} support this interpretation: removing context-clean corruption or span-aware attention alignment lowers accuracy (15.5\% $\rightarrow$ 14.5\%) and increases invalid actions (up to 7.1\%), indicating that structure-aware training components partially mitigate span-level formatting violations while preserving planning efficiency. Removing either masking component reduces overall accuracy by about 1\% (15.5\% $\rightarrow$ 14.5\%), indicating that training--inference alignment in masking modestly improves performance under the same budget.
In addition, removing span-aware attention alignment increases the invalid action rate (6.4\% $\rightarrow$ 7.1\%), suggesting that span-level structural alignment contributes to syntactic reliability in tool-use trajectories.

\paragraph{Inductive biases under matched success rates.}
\zhenew{Although the two paradigms achieve identical average accuracy, they tend to succeed on different subsets of questions due to distinct inductive biases.
AR decoding commits token-by-token, which favors strict structured outputs (e.g., tool-call schemas, JSON formatting, delimiter balancing) and incremental retrieval--verification loops with frequent local checks.
In contrast, diffusion-style generation refines action spans globally, which favors multi-constraint coordination, contract-driven planning, and long-form consistency across steps.
This architectural difference explains how DLLM can match AR in final accuracy while using fewer interactions: global span refinement encourages early plan formation and reduces iterative backtracking, but may occasionally sacrifice local syntactic precision without additional structural constraints.}

\paragraph{Calibration and reliability trade-offs.}
Despite matching AR in final-answer correctness, DLLM exhibits worse calibration (Calibration Error: 66\% vs.\ 59\%; lower is better). This indicates that its episode-level confidence is less aligned with empirical correctness.
One plausible explanation is trajectory compactness: by committing to fewer intermediate verification steps, DLLM has fewer opportunities to update uncertainty estimates, potentially leading to over-confident failures on difficult episodes.
The higher invalid action rate (6.4\% vs.\ 1.9\%) further contributes to this gap, as local schema violations—rather than reasoning errors—can decouple internal confidence from realized correctness and inflate ECE. 
The ablations follow the same trend: removing context-clean corruption or span-aware attention alignment reduces accuracy (to 14.5\%) and worsens calibration (ECE increases to 67\%--68\%), suggesting that structure-aware components support not only efficiency and planning coverage but also more reliable confidence estimates.
Overall, these results highlight a practical efficiency--reliability trade-off: diffusion backbones can reduce interaction cost under matched accuracy, but may require additional calibration- and constraint-aware techniques to make confidence scores actionable for risk-sensitive tool use (e.g., deciding when to verify, abstain, or allocate additional budget).

\paragraph{Invalid action rate versus final accuracy.}
The invalid-action metric is episode-level: an episode is marked invalid if it contains \emph{any} unparsable span, even if later turns recover with a valid tool call or terminal answer.
Therefore, a higher invalid-action rate does not necessarily translate into a proportional drop in final accuracy, since malformed tool calls can often be corrected in subsequent turns.
We attribute the gap primarily to generation dynamics: diffusion refines spans globally and may occasionally violate strict formatting constraints, whereas AR benefits from token-level conditioning that better preserves local syntactic structure.
This observation motivates future work on tool-call-oriented supervision and structure-aware decoding for diffusion-based agents in multi-turn tool-use settings.

\subsection{\zhenew{Trajectory-Level Behavioral Differences}}

Beyond aggregate accuracy and efficiency metrics, we examine whether the two paradigms exhibit systematically different decision dynamics at the trajectory level.
First, DLLM produces more compact interaction sequences, using fewer tool calls (6.7 vs.\ 7.5) and fewer turns (13.0 vs.\ 14.8), resulting in over 30\% reduction in end-to-end inference time under matched budgets. This suggests faster convergence rather than merely reduced token generation.
Second, planning statistics reveal a structural difference in decision allocation. DLLM exhibits substantially higher planning coverage (PlanRate: 62\% vs.\ 38\%; Plan@Ep: 0.92 vs.\ 0.55), indicating more frequent explicit global plan formation prior to tool execution. In contrast, AR decoding proceeds incrementally, committing left-to-right and revising locally when conflicts arise.

\begin{figure}[htbp]
    \centering
    \subfigure[Backtracking comparison]{\includegraphics[width=0.49\linewidth]{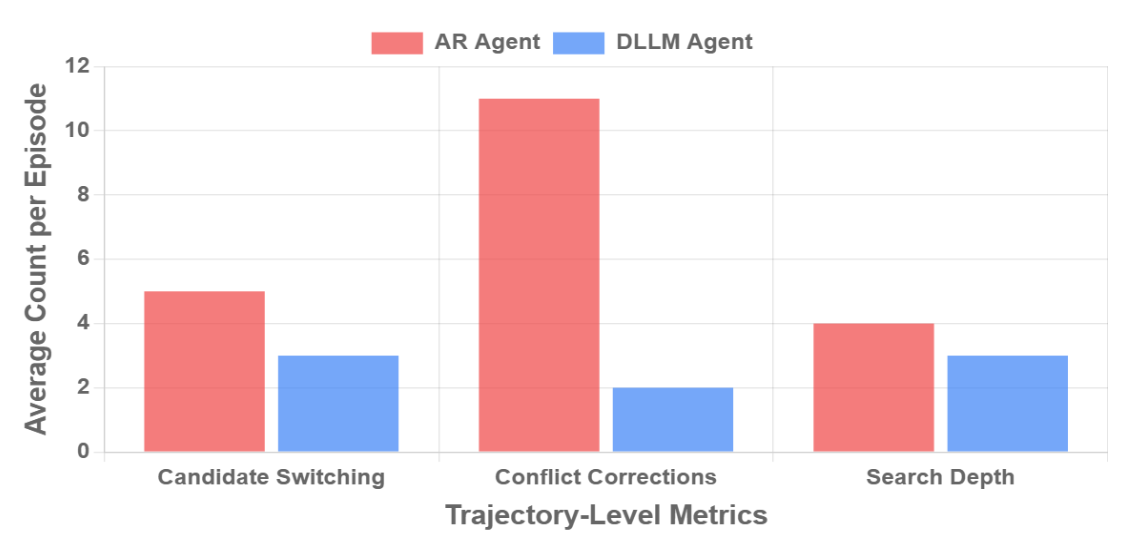}}
    \subfigure[Detailed accuracy comparison]{\includegraphics[width=0.46\linewidth]{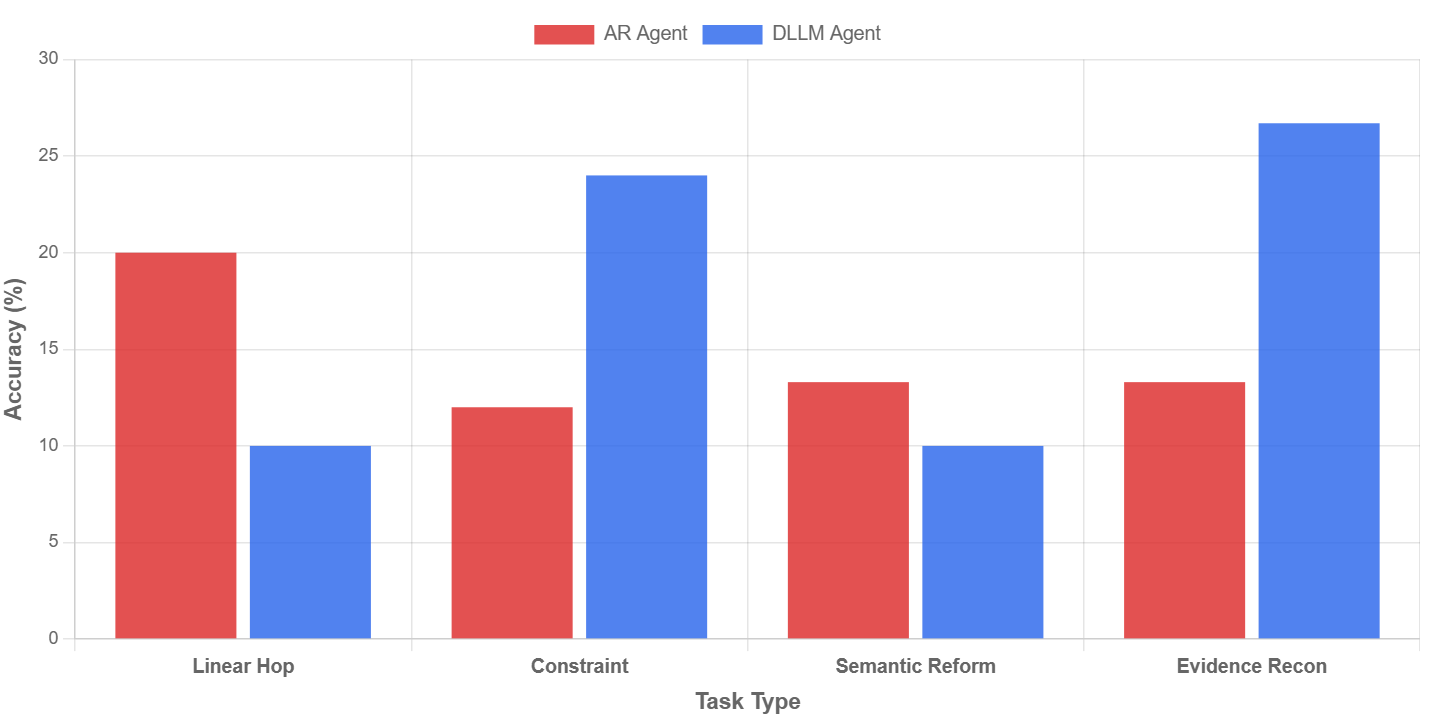}}
    \caption{(a) Accuracy-efficiency trade-off.  Each point represents a distinct inference strategy, plotted by average inference time (seconds) and task accuracy (\%). Inference time is averaged per episode under identical hardware and tool API settings. All methods share the same backbone size, tool interface, action parser, and interaction budget; the only difference lies in the inference strategy (standard AR and DLLM agent, and with BON, beam search, or Tree-of-Thought.). (b) Accuracy breakdown across four structural task categories. Although the two paradigms achieve identical overall accuracy (15.5\%), their strengths distribute differently across task structures. AR performs better on linear multi-hop chains, while DLLM shows clear advantages on constraint intersection and evidence reconciliation tasks, suggesting complementary structural inductive biases.
}
    \label{fig:backtrack_com}
\end{figure}

To further quantify trajectory stability, Figure~\ref{fig:backtrack_com}(a) reports three turn-level structural metrics averaged across episodes where \emph{both} AR and DLLM agents successfully reach the correct answer. This controlled subset isolates trajectory dynamics independent of final accuracy, ensuring that differences are not confounded by failure cases. 
Within this matched-success set, AR agents exhibit substantially higher trajectory instability, with more candidate switching (5 vs.\ 3 on average) and dramatically more conflict corrections (11 vs.\ 2), indicating frequent local revisions after committing to intermediate hypotheses. While the average search depth difference is smaller (4 vs.\ 3), the much larger gap in conflict correction events suggests that AR trajectories are characterized less by deeper exploration and more by repeated backtracking cycles. In contrast, DLLM trajectories involve fewer hypothesis shifts and significantly fewer conflict-driven revisions, reflecting earlier global alignment across constraints even when both paradigms ultimately produce correct answers.

Figure \ref{fig:backtrack_com}(b) presents the accuracy breakdown across four structural categories. Although AR and DLLM achieve identical overall accuracy (15.5\%), their performance differs substantially across task structures. On linear multi-hop chains, AR outperforms DLLM (20.0\% vs.\ 10.0\%), indicating that sequential left-to-right commitment is well-suited for dependency chains where reasoning follows a relatively linear progression. In contrast, DLLM demonstrates clear advantages on constraint intersection (24.0\% vs.\ 12.0\%) and evidence reconciliation tasks (26.7\% vs.\ 13.3\%), both of which require coordinating multiple constraints or resolving conflicting evidence. Performance on semantic reformulation tasks is comparable (13.3\% vs.\ 10.0\%), suggesting that neither paradigm holds a decisive advantage when lexical obfuscation dominates structural complexity.
These results indicate that the two paradigms exhibit complementary strengths: autoregressive decoding benefits linear dependency tracking, while diffusion-style generation better supports global constraint coordination and conflict resolution.

Finally, qualitative inspection of complex evidence-reconciliation cases aligns with these aggregate statistics. AR agents often exhibit candidate switching and late conflict detection, sometimes failing to converge within 15 turns (>1600s), whereas DLLM converges within 6 turns (418s) by aligning multiple constraints early. Taken together, both quantitative trajectory metrics and case-level analysis indicate a systematic paradigm difference: AR decoding favors incremental exploration with frequent local corrections, whereas diffusion-style generation encourages earlier global coordination and reduced backtracking.

\subsection{\zhenew{Detailed Analysis}}

\zhenew{BrowseComp-zh \cite{zhou2025browsecomp} is deliberately constructed to stress-test long-horizon web reasoning under constrained search conditions. Several core properties distinguish it from standard multi-hop QA benchmarks. 
First, many questions are \emph{reverse-constructed} from known answers, increasing compositional difficulty and reducing shortcut patterns. Second, problems typically involve \emph{multi-constraint composition}, where temporal, spatial, categorical, or descriptive conditions must be jointly satisfied. Third, answers are intentionally \emph{non-retrievable from the first search result page}, forcing deeper browsing and query reformulation. Fourth, relevant evidence is often \emph{distributed across heterogeneous platforms}, requiring cross-source alignment and conflict resolution rather than simple snippet extraction. From the perspective of reasoning structure, we categorize BrowseComp-zh questions into four major types}:

\subsubsection*{Linear Multi-hop Chain}

Linear multi-hop chain questions require sequential evidence accumulation across multiple documents or pages, where each intermediate finding directly enables the next retrieval step. The original benchmark description explicitly emphasizes ``multi-step reasoning,'' ``multi-hop retrieval,'' and ``cross-page reasoning'' as central challenges \cite{zhou2025browsecomp}. The reasoning trajectory therefore forms a linear dependency chain, and failure to resolve any intermediate hop typically leads to an incorrect final answer. 

Empirically, \zhenew{AR agent demonstrate stronger performance than DLLM agent on this category}. 
Because the reasoning structure is explicitly sequential, the AR paradigm aligns naturally with the task dynamics: each action is conditioned on previously retrieved evidence, and explicit inter-turn search branching allows the agent to progressively extend the reasoning chain. 
The task difficulty primarily lies in depth rather than combinatorial constraint interaction, making step-by-step decomposition both sufficient and reliable. In contrast, DLLM agent rely on intra-step refinement and global latent planning. 
While such global coordination can be advantageous in more structurally coupled tasks, it does not provide a clear benefit in strictly linear chains. As a result, for problems dominated by sequential evidence accumulation rather than constraint coupling, inter-turn AR search provides a more suitable inductive bias.

Although this case may appear to involve multiple constraints at first glance, its structural nature differs fundamentally from constraint-intersection problems: (1) The question contains multiple descriptive elements, but they do not function as independent filtering axes. (2) There are no simultaneously enforced conditions that must be jointly satisfied. (3) Nor does the reasoning process require cross-branch filtering or combinatorial intersection across orthogonal constraints. Instead, the structure forms a \textit{strict linear dependency chain} centered around a single pivot variable—the scientist’s birth year. Each subsequent condition is deterministically derived from the previous step, creating a unidirectional sequence of entity-time mappings. The apparent complexity arises from chain depth rather than constraint coupling. 
Under such structural conditions, the AR paradigm aligns directly with the task’s sequential dependency pattern. By contrast, diffusion-style planning emphasizes intra-step global semantic alignment and holistic temporal modeling, and it offers no structural benefit when reasoning unfolds along a single linear axis. Therefore, despite its surface appearance of multi-constraint reasoning, this case structurally favors inter-turn AR expansion over intra-step global refinement. The following case shows the comparison between AR and DLLM agents on solving linear multi-hop chain reasoning problem.

\begin{tcolorbox}[
    colback=gray!10,
    colframe=gray!60,
    title={Case Study: Linear Multi-hop Chain Reasoning Pattern},
    fonttitle=\bfseries,
    breakable
]

\textbf{Problem (English Translation).}
\smallskip
\emph{
A medical scientist studied abroad in the early 1900s. In the following year, he founded a journal promoting patriotic spirit, which gained notable influence. 
He was highly praised by a top university in his home province. 
In the year he was born, a neighboring country of China welcomed a new head of government. 
The teacher of that head of government was a Chinese philosopher from a particular intellectual school, who died at a relatively young age. 
In the year of the philosopher’s death, China became the third country in the world to master a certain technology. 
Who is this medical scientist?
}

\bigskip
\textbf{Structural Property.}  
This question forms a strict linear dependency chain centered on a single pivot variable (the scientist’s birth year). 
Each step enables the next retrieval, without requiring multi-constraint intersection or cross-branch reconciliation.

\bigskip

\begin{minipage}[t]{0.48\textwidth}
\textbf{AR Agent}
\medskip

\textit{Reasoning Style: Explicit Step-by-Step Chain Expansion}

\medskip
\begin{itemize}[leftmargin=*, itemsep=3pt]
\item Search: ``medical scientist 1900s studied abroad founded journal patriotism'' $\rightarrow$ Candidate A
\item Search: ``Candidate A birth year'' $\rightarrow$ 1885
\item Search: ``Neighboring country head of government in 1885'' $\rightarrow$ X
\item Search: ``Teacher of X'' $\rightarrow$ Y
\item Search: ``Year of Y's death'' $\rightarrow$ 19xx
\item Search: ``China third country master technology 19xx'' $\rightarrow$ Verification
\end{itemize}

\end{minipage}
\hfill
\begin{minipage}[t]{0.48\textwidth}

\textbf{DLLM Agent}

\medskip

\textit{Reasoning Style: Global Latent Planning Before Execution}

\medskip

\begin{itemize}[leftmargin=*, itemsep=3pt]
\item Planning phase identifies:
    \begin{itemize}
    \item This is a temporal dependency chain
    \item Core pivot variable: birth year
    \item Required mapping:
    
    Scientist $\rightarrow$ Birth Year  
    $\rightarrow$ Foreign Head of Government  
    $\rightarrow$ Teacher  
    $\rightarrow$ Death Year  
    $\rightarrow$ Technology Milestone
    \end{itemize}
\item Forms a globally aligned temporal representation
\item Decides retrieval order after latent refinement 
\item Searching loop under random mask 
\end{itemize}

\end{minipage}
\end{tcolorbox}

\subsubsection*{Constraint Intersection}

Constraint-intersection questions involve simultaneously satisfying multiple orthogonal constraints, such as temporal, spatial, categorical, and descriptive filters. The benchmark explicitly highlights a ``multi-constraint design: temporal, spatial, categorical, descriptive conditions combined,'' indicating that correct answers must satisfy several independent conditions at once. Unlike linear chains, the core difficulty arises from combinatorial filtering and precise condition tracking rather than reasoning depth alone.

\zhenew{DLLM agent outperforms AR agent in this category}. The advantage appears to stem from the diffusion-style planner's global refinement mechanism. Rather than committing early to a single retrieval branch, the DLLM agent forms a more holistic semantic alignment across constraints before issuing tool calls. This reduces premature commitment and mitigates local confirmation bias that often emerges in AR stepwise exploration. However, AR agent tend to explore constraints incrementally. While this strategy works well for linear chains, it increases the risk of branching into locally plausible yet globally inconsistent paths when constraints are strongly coupled. The need for simultaneous constraint satisfaction therefore structurally favors planning mechanisms that support global alignment, explaining the observed performance gap.

To further examine the interaction between reasoning structure and base model architecture, we construct a multi-constraint example as a controlled structural modification of the preceding linear multi-hop chain case. Rather than introducing a new domain or increasing task difficulty, we retain comparable semantic content while altering the underlying reasoning topology. Specifically, the original linear multi-hop chain is reformulated into a constraint-intersection problem in which multiple independent conditions must be satisfied simultaneously. This controlled transformation enables us to isolate how different agent paradigms respond when the structural organization of the task shifts from sequential dependency depth to multi-axis constraint coupling.

In the new setting, the reasoning objective is defined by the joint satisfaction of multiple orthogonal conditions rather than by a single sequential dependency chain. The AR agent approaches the task through incremental constraint accumulation, verifying one attribute at a time and progressively narrowing the candidate set. However, this stepwise expansion strategy often leads to early commitment to partially compatible candidates. Because independent constraints are not globally aligned at each stage, inconsistencies tend to be discovered only after several additional retrieval steps. As a result, the agent frequently engages in backtracking and re-querying cycles, revisiting earlier branches to repair local conflicts. Although the final answer is eventually correct, this fragmented exploration process introduces substantial redundancy. Empirically, the AR agent required 18 interaction turns and 1219 seconds to complete the task, reflecting high retrieval redundancy and elevated search entropy. In contrast, the DLLM agent performs simultaneous multi-constraint modeling during its planning phase. Instead of sequentially committing to partial candidates, it encodes all constraints into a globally aligned latent representation and evaluates candidate compatibility across dimensions in parallel. This reduces premature commitment and minimizes the need for corrective backtracking. The search process therefore exhibits a more structured and parallel retrieval loop, with fewer redundant queries. In our experiment, the DLLM agent completed the task in 6 interaction turns and 419 seconds, demonstrating substantially lower redundancy and improved search efficiency.

\begin{tcolorbox}[
    colback=gray!10,
    colframe=gray!60,
    title={Case Study: Constraint-Intersection Reasoning Pattern},
    fonttitle=\bfseries,
    breakable
]

\textbf{Problem (English Translation).}
\smallskip
\emph{
A medical scientist satisfies the following conditions:
(1) He studied abroad in the early 1900s and founded a patriotic medical journal the following year.
(2) He later served at a top university in a coastal province of China.
(3) He was born in the late Qing dynasty.
(4) One of his students later became a key public health figure during wartime.
(5) He participated in drafting an important early modern Chinese public health regulation.
Who is this medical scientist?
}

\bigskip
\textbf{Structural Property.}  
Unlike a linear dependency chain, this problem requires the simultaneous satisfaction of multiple independent constraints 
(spanning temporal, institutional, relational, and historical dimensions). 
The correct answer lies at the intersection of these orthogonal filtering axes. 
No single pivot variable deterministically generates the others.

\bigskip

\begin{minipage}[t]{0.48\textwidth}

\textbf{AR Agent}

\medskip
\textit{Reasoning Style: Sequential Constraint Accumulation with Backtracking}

\medskip

\begin{itemize}[leftmargin=*, itemsep=3pt]
\item Turn 1–3: Search based on overseas study + journal founding $\rightarrow$ Candidate Set A
\item Turn 4–6: Apply university constraint $\rightarrow$ Narrow to Candidate B
\item Turn 7–9: Apply student wartime role constraint $\rightarrow$ Mismatch detected
\item Turn 10–12: Backtrack to earlier candidate pool
\item Turn 13–15: Re-search institutional affiliation details
\item Turn 16–18: Verify public health legislation participation $\rightarrow$ Final Answer
\end{itemize}

\end{minipage}
\hfill
\begin{minipage}[t]{0.48\textwidth}

\textbf{DLLM Agent}

\medskip
\textit{Reasoning Style: Parallel Constraint Encoding with Global Alignment}

\medskip

\begin{itemize}[leftmargin=*, itemsep=3pt]
\item Turn 1: Encode all constraints jointly in planning phase
\item Turn 2–3: Parallel retrieval targeting multi-attribute matching
\item Turn 4: Cross-condition consistency verification
\item Turn 5–6: Final confirmation and answer emission
\end{itemize}

\end{minipage} 
\end{tcolorbox}

\subsubsection*{Implicit Semantic Reformulation with Lexical Obfuscation}

BrowseComp-zh explicitly disallows answers from appearing on any search engine’s first result page, frequently inducing lexical obfuscation and indirect phrasing. 
As a consequence, surface-form keyword matching is insufficient; agents must perform query reformulation and semantic rewriting to bridge lexical gaps. 
This design intentionally suppresses direct keyword shortcuts and requires abstraction beyond literal query tokens.

Empirically, this category exhibits a different pattern. Under relaxed settings—specifically, \zhenew{when the number of interaction turns is unrestricted and the context window extends to 128k tokens—AR agent achieve higher accuracy}. 
With extended context and interaction capacity, AR agent can iteratively refine queries, accumulate evidence, and gradually correct lexical mismatches. The strength of inter-turn reformulation becomes more pronounced when memory constraints are removed. However, \zhenew{under a 32k context limitation, AR agent lose this advantage}. 
Restricted context constrains long-horizon query rewriting and evidence tracking, both of which are essential for overcoming lexical suppression. 
DLLM agent, during planning, tend to form an internal global semantic alignment that partially compensates for lexical mismatch even under tighter memory budgets. Nevertheless, this potential advantage is offset by structured tool-calling instability: formatting errors and invocation inconsistencies prevent the diffusion-based planner from consistently translating semantic alignment into successful retrieval trajectories. Consequently, neither paradigm exhibits a decisive advantage under strict context constraints.

\begin{tcolorbox}[
    colback=gray!10,
    colframe=gray!60,
    title={Case Study: Implicit Semantic Reformulation with Lexical Obfuscation},
    fonttitle=\bfseries,
    breakable
]

\textbf{Problem (English Translation).}
\smallskip
\emph{
In recent years, a popular folk-style song was released. 
The same individual composed the lyrics, wrote the music, and performed the song. 
The arrangement incorporates suona and erhu elements, conveying a strong sense of historical depth and cultural heritage. 
In the year of its release, the song ranked within the Top 5 of the ``YouNiBang Annual Chinese-style Songs'' chart. 
Question: Where is this all-round singer’s most well-known female disciple from?
}

\bigskip
\textbf{Structural Property.}  
This problem does not primarily require multi-hop dependency chaining or orthogonal constraint intersection. 
Instead, it hinges on implicit semantic reformulation. 
The core entity (a specific song) is described through culturally loaded and paraphrastic expressions rather than canonical identifiers (e.g., title or artist name). 
Successful retrieval depends on compressing dispersed lexical cues into a standardized query representation before entity resolution.

\bigskip

\begin{minipage}[t]{0.48\textwidth}

\textbf{AR Agent}

\medskip
\textit{Reasoning Style: Direct Surface-Form Retrieval with Iterative Refinement}

\medskip

\begin{itemize}[leftmargin=*, itemsep=3pt]
\item Turn 1–4: Directly search using original description (``folk song'', ``suona'', ``erhu'', ``historical depth'') $\rightarrow$ Multiple candidate songs
\item Turn 5–8: Attempt refinement using ranking clue (``YouNiBang Top 5'') $\rightarrow$ Still ambiguous matches
\item Turn 9–14: Backtrack and modify query combinations $\rightarrow$ Partial overlaps but no unique identification
\item Turn 15: \textbf{Still no confirmed target entity}
\item Turn 16–24: Reconstruct query around chart year + composer-performer identity
\item Turn 25–31: Narrow to single artist
\item Turn 32: Identify most well-known female disciple $\rightarrow$ Retrieve birthplace (Final Answer)
\end{itemize}

\end{minipage}
\hfill
\begin{minipage}[t]{0.48\textwidth}

\textbf{DLLM Agent}

\medskip
\textit{Reasoning Style: Latent Semantic Compression with Canonical Query Generation}

\medskip

\begin{itemize}[leftmargin=*, itemsep=3pt]
\item Turn 1: Encode full description jointly in latent space
\item Turn 2: Compress dispersed cues into canonical form  
      (``specific YouNiBang Top 5 Chinese-style song by singer-songwriter incorporating traditional instruments'')
\item Turn 3–6: Generate standardized retrieval query targeting chart records
\item Turn 7–10: Correctly identify target song and singer (planner accurate)
\item Turn 11–15: Tool invocation errors during disciple lookup $\rightarrow$ Looping behavior
\item Exceeds predefined context limit $\rightarrow$ Terminated without final answer
\end{itemize}

\end{minipage}
\end{tcolorbox}

The following case shows the difference between AR and DLLM agents for the case of lexical obfuscation. 
In this task, the core challenge lies in semantic compression rather than constraint intersection or dependency chaining. The target entity is described through dispersed, culturally loaded expressions instead of canonical identifiers, requiring the agent to reconstruct a standardized representation before effective retrieval can occur. The AR agent begins by directly querying the original surface description. However, because the lexical cues are fragmented and loosely aligned with standard knowledge base entries, the initial searches return multiple partially overlapping candidates. This results in a high branching factor and repeated local commitments to incorrect entities. Within the first 15 interaction turns, the AR agent fails to uniquely identify the target song, as it repeatedly backtracks and reformulates queries in response to inconsistencies. Only after progressive consolidation of the lexical clues does it converge to the correct entity at Turn 32, ultimately retrieving the correct answer after 3820 seconds. Although successful, the process exhibits substantial redundancy, heavy reliance on surface lexical overlap, and late-stage semantic consolidation.

By contrast, the DLLM agent performs latent semantic alignment at the planning stage. Instead of issuing fragmented keyword queries, it compresses the dispersed cues into a globally coherent semantic representation and generates a more canonical retrieval query. This significantly reduces lexical search entropy and yields a structurally accurate plan for identifying the target song and singer. However, despite correct planning and intermediate entity identification, the agent encounters instability in tool invocation during the disciple lookup phase. The resulting execution loop prevents it from producing a final answer within 15 turns, and the process terminates after exceeding the predefined context budget. Thus, while the DLLM agent demonstrates superior semantic reformulation capability, this case reveals that end-to-end success in implicit reformulation tasks depends not only on global alignment but also on robust tool orchestration. A detailed step-by-step comparison is provided in the following case box.

\subsubsection*{Evidence Reconciliation}

Evidence-reconciliation questions involve integrating fragmented, and sometimes partially conflicting, information scattered across heterogeneous sources. 
The benchmark explicitly notes that ``effectively reconciling retrieved evidence with internal representations remains non-trivial,'' underscoring the importance of cross-source alignment and contradiction resolution. 
These tasks require not only retrieval but also coherent integration of heterogeneous evidence into a consistent final conclusion.

\zhenew{In this category, DLLM agent again demonstrate superior performance. } 
The diffusion-based planner benefits from intra-step global refinement, enabling it to maintain a more coherent latent representation of accumulated evidence before committing to a final answer. 
This global coordination reduces over-reliance on early retrieved snippets and mitigates confirmation bias.
AR agent, in contrast, are more susceptible to anchoring effects due to their sequential commitment mechanism. 
If early retrieval results are incomplete or misleading, explicit backtracking becomes necessary, increasing the likelihood of failure. Thus, tasks that demand cross-source reconciliation and global consistency checking appear structurally aligned with the strengths of diffusion-based planning. 

The following case compares the DLLM and AR agents on evidence reconciliation. 
In this task, the primary difficulty lies in aligning multiple narrative-level clues that are individually non-discriminative but collectively decisive. The AR agent exhibits a typical early-commitment pattern: it anchors on a high-probability candidate based on salient traits such as “unrestrained spirit” and “romantic works,” and then sequentially verifies subsequent evidence. When later clues—particularly the modern song reference—introduce inconsistencies, the agent is forced to backtrack and test alternative figures. This repeated cycle of commitment, conflict detection, and candidate switching prevents convergence within the first 15 turns. Notably, even after more than 1600 seconds of execution time, the AR agent still fails to produce a confirmed answer, reflecting the inefficiency induced by sequential evidence validation and local revision.

By contrast, the DLLM agent encodes personality traits, biographical signals, literary style, and modern cultural reference jointly in latent space before committing to any single candidate. This global evidence integration enables it to construct a unified representation against which candidates are aligned in parallel, avoiding premature commitment and reducing reconciliation overhead. As a result, the DLLM agent identifies the uniquely consistent figure within 6 turns, with an end-to-end execution time of 418 seconds. The temporal contrast highlights a structural advantage: in evidence-reconciliation problems, global semantic alignment substantially reduces iterative backtracking and accelerates convergence, whereas sequential verification strategies are prone to compounding delays under narrative ambiguity.

\begin{tcolorbox}[
    colback=gray!10,
    colframe=gray!60,
    title={Case Study: Evidence Reconciliation},
    fonttitle=\bfseries,
    breakable
]

\textbf{Problem (English Translation).}
\smallskip
\emph{
In China, there was a literary creator known for his “unrestrained” spirit. 
In his youth, due to his remarkable talent, he received a special “summons.” 
This summons led him to travel widely, making him a subject of public fascination. 
He created many works imbued with romantic temperament, whose artistic imagery has been deeply appreciated by later generations. 
Recently, a song named after him once again brought his name into the public spotlight. 
Who is this free-spirited creator?
}

\bigskip
\textbf{Structural Property.}  
This problem requires reconciling multiple pieces of narrative-level evidence spanning personality traits, biographical experiences, literary style, and modern cultural references. 
Each individual clue is non-unique and potentially compatible with multiple historical figures. 
The solution emerges only after aligning all evidence fragments into a globally consistent candidate representation and resolving conflicts among competing interpretations.

\bigskip

\begin{minipage}[t]{0.48\textwidth}

\textbf{Autoregressive (AR) Agent}

\medskip
\textit{Reasoning Style: Early Candidate Commitment with Sequential Verification}

\medskip

\begin{itemize}[leftmargin=*, itemsep=3pt]
\item Turn 1–3: Retrieve candidates based on ``unrestrained'' + ``romantic works'' $\rightarrow$ Early commitment to a high-probability figure (e.g., Li Bai)
\item Turn 4–6: Validate against ``special summons'' and ``travel widely'' clues $\rightarrow$ Partial alignment
\item Turn 7–9: Introduce modern song reference $\rightarrow$ Inconsistency detected
\item Turn 10–12: Backtrack and test alternative candidates
\item Turn 13–15: Further conflict across biographical and stylistic evidence $\rightarrow$ \textbf{No confirmed answer within 15 turns}
\item Subsequent turns: Repeated candidate switching and verification cycles
\end{itemize}

\end{minipage}
\hfill
\begin{minipage}[t]{0.48\textwidth}

\textbf{Diffusion-based LLM (DLLM) Agent}

\medskip
\textit{Reasoning Style: Global Evidence Encoding with Candidate Alignment}

\medskip

\begin{itemize}[leftmargin=*, itemsep=3pt]
\item Turn 1: Jointly encode personality, biography, literary style, and modern reference in latent space
\item Turn 2–3: Form global candidate embedding representing all constraints simultaneously
\item Turn 4–5: Align candidate set against unified representation
\item Turn 6: Identify uniquely consistent figure $\rightarrow$ Final Answer
\end{itemize}

\end{minipage}
\end{tcolorbox}

\subsection{Performance–Efficiency Trade-offs Across Inference Strategies}

\begin{figure}[htbp]
    \centering
    \includegraphics[width=0.7\linewidth]{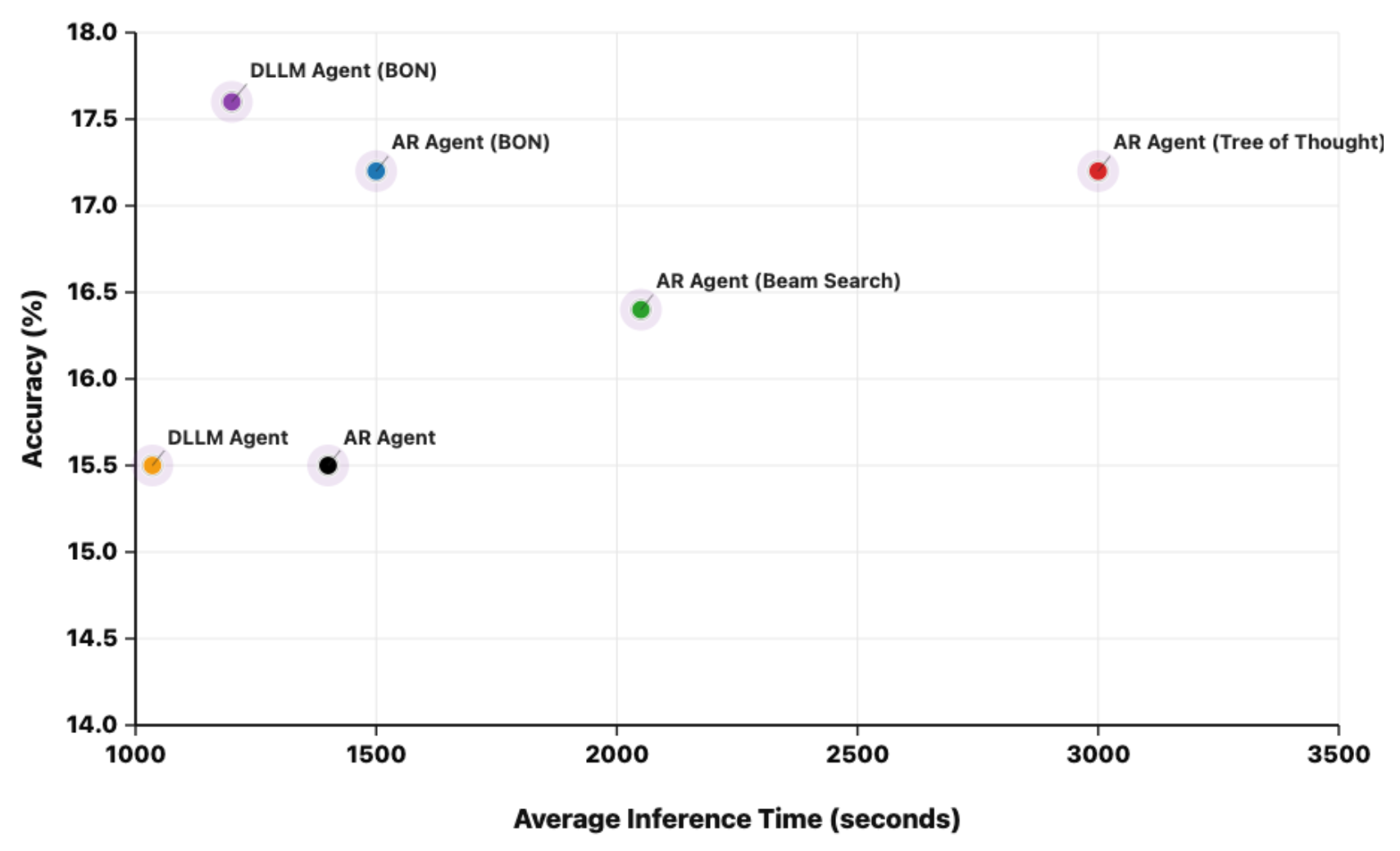}
    \caption{Accuracy-efficiency trade-off.  Each point represents a distinct inference strategy, plotted by average inference time (seconds) and task accuracy (\%). Inference time is averaged per episode under identical hardware and tool API settings. All methods share the same backbone size, tool interface, action parser, and interaction budget; the only difference lies in the inference strategy (standard AR and DLLM agent, and with BON, beam search, or Tree-of-Thought.).
}
    \label{fig:tradeoff}
\end{figure}

\zhenew{Figure~\ref{fig:tradeoff} illustrates the performance–efficiency trade-off across different inference strategies.
Search-based AR variants, including Best-of-$N$ (BON), Beam Search, and Tree-of-Thought (ToT), provide modest accuracy improvements over vanilla AR decoding.
However, these gains come at a substantial computational cost.
In particular, ToT increases inference latency by nearly $3\times$ relative to DLLM, while yielding only marginal improvements in accuracy.
Beam search exhibits a similar pattern, incurring significantly higher latency without commensurate performance gains.
In contrast, DLLM achieves accuracy comparable to standard AR decoding while reducing inference time by approximately $32\%$.
Relative to AR-based BON, Beam Search, and ToT, DLLM reduces latency by approximately $40\%$, $92\%$, and $66\%$, respectively.
These results position DLLM in a favorable region of the Pareto frontier, demonstrating that structured latent planning improves computational efficiency without sacrificing task performance.}

\zhenew{We further investigate whether search-based frameworks can provide additional benefits on top of DLLM.
To this end, we implement a DLLM-based BON variant following the same sampling protocol used for AR agent.
Compared to the base DLLM strategy, DLLM with BON increases inference time by approximately $18\%$, while delivering consistent accuracy gains.
Notably, despite the additional sampling cost, DLLM with BON achieves slightly higher accuracy than AR with BON.
This suggests that improving planning quality at the architectural level provides a stronger foundation for downstream search than relying solely on decoding-time exploration.}

Overall, these results indicate that architectural improvements in planning (DLLM) are significantly more computationally efficient than heavy decoding-time search, while lightweight sampling on top of DLLM offers a flexible mechanism to further improve performance when additional latency is acceptable.

%% file: arxiv/analysis.tex
\section{Analysis from Confidence and Entropy Evolution}  \label{sec:analysis}

While Section \ref{sec:exp} demonstrates that DLLM agents achieve substantial efficiency gains—particularly the ability to complete tasks with fewer tool invocations and shorter trajectories—the underlying mechanisms driving these workflow-level differences remain unclear. Are these advantages purely attributable to token-level parallelism, or do diffusion-style generation patterns manifest in qualitatively different planning and tool-use behaviors? To answer this question, we analyze the internal decoding dynamics of DLLM agents during multi-step reasoning workflows. Through fine-grained examination of token-level decoding order, confidence patterns, and entropy evolution within the multi-agent workflow, we reveal how the diffusion backbone shapes agent behavior at a structural level. We focus on two critical roles—the Planner Agent (responsible for task decomposition) and the Information Seeker Agent (handling information retrieval)—to illustrate paradigm-specific behavioral patterns. To make these patterns concrete, we conduct a detailed case study on the following query:

\begin{compactcase}
A tech company: its name comes from an animal; its parent company is a Chinese internet company; it was formed by merging two in-house teams; and it involves both software and hardware. What is the company called?
\end{compactcase}

\begin{figure}[htbp]
    \centering
    \includegraphics[width=0.85\linewidth]{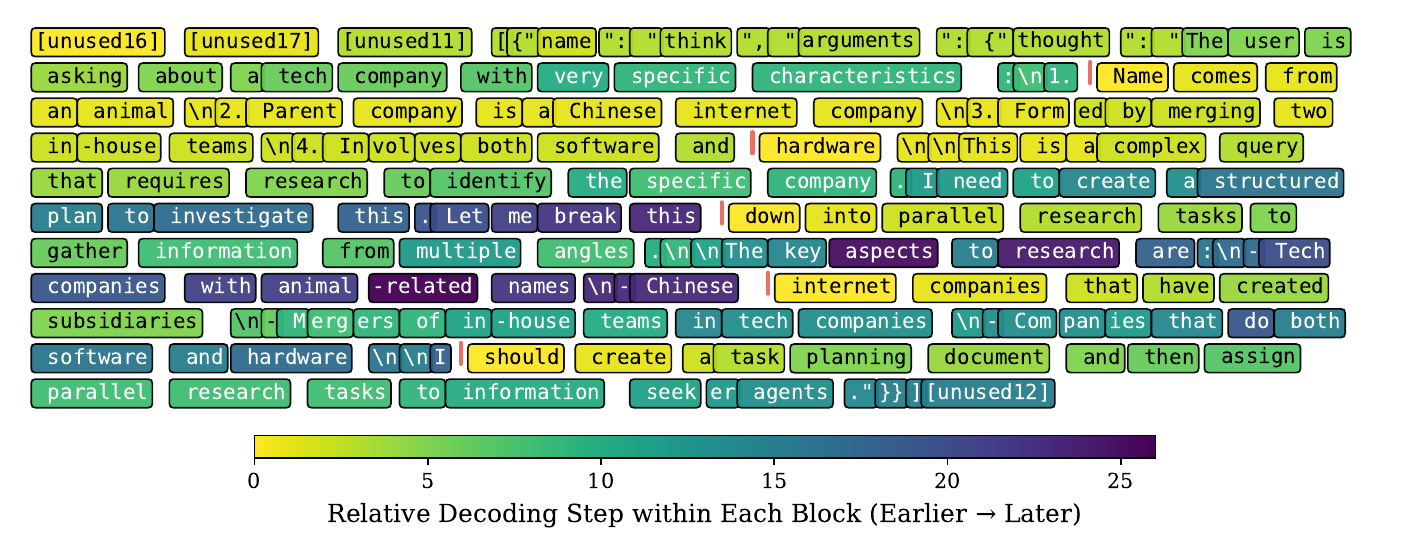}
    \caption{Heatmap showing the relative decoding order of each token within each block of an example Planner Agent response (block size = 32). Different blocks are separated by red lines.}
    \label{fig:planner_heatmap}
\end{figure}

\begin{figure}[htbp]
    \centering
    \includegraphics[width=1\linewidth]{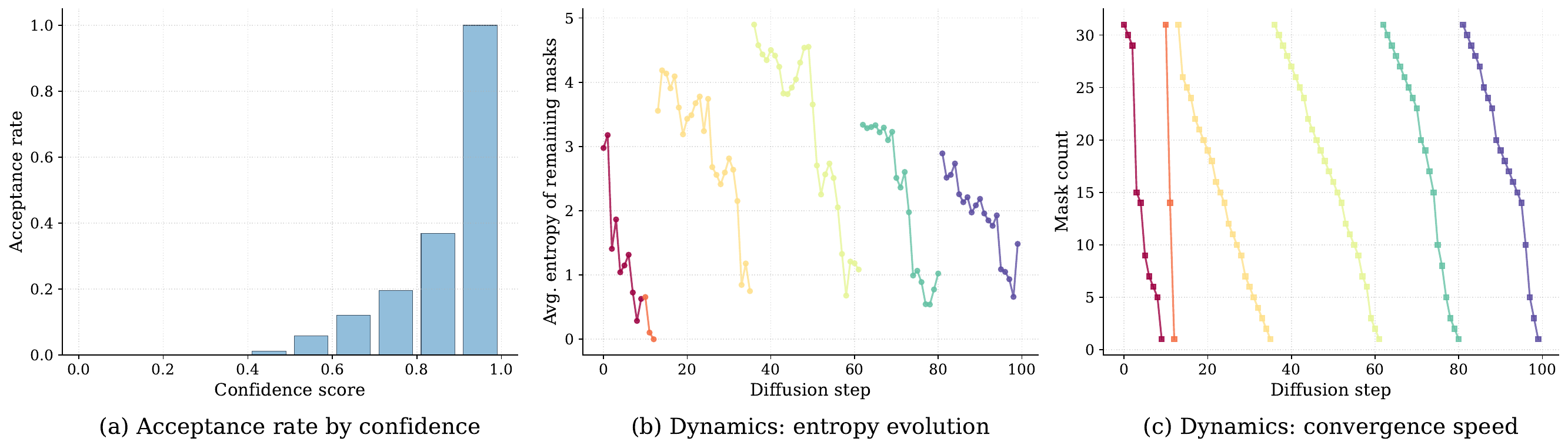}
    \caption{Analysis of the Planner Agent response shown in Figure~\ref{fig:planner_heatmap}. (a) Probability of tokens with different confidence scores being decoded during diffusion process. (b) Average entropy of remaining mask tokens across different diffusion steps. (c) Number of remaining mask tokens within each block across different diffusion steps. In (b) and (c), different colors represent the decoding process of different blocks.}
    \label{fig:planner_chart}
\end{figure}

\subsection{Planner Agent}

The Planner Agent operates at the workflow's inception, tasked with decomposing complex user queries into actionable research tasks. Figure~\ref{fig:planner_heatmap} presents an example response from the Planner Agent, with its heatmap revealing the relative decoding order of each token within each block. Meanwhile, Figure~\ref{fig:planner_chart} illustrates the underlying decoding dynamics of generating this response. As shown in Figure~\ref{fig:planner_heatmap}, the response can be divided into two distinct phases: (1) understanding and summarizing the user's request (corresponding to blocks 1-2), and (2) performing task decomposition (corresponding to blocks 3-6). The following analysis examines the decoding dynamics of these two phases.

\textbf{Phase 1: Understanding and Summarization (Blocks 1-2).}
During this initial phase, which covers the first two blocks of the response, we observe that the planner agent can rapidly extract the four key pieces of information from the user query. Specifically, the first and second key points are decoded almost simultaneously within the same diffusion step, while the third and fourth key points are decoded together in another diffusion step. This behavior demonstrates that DLLM possesses strong information summarization and extraction capabilities, enabling it to rapidly output key points in parallel. As shown in Figure~\ref{fig:planner_chart}(a), the high parallelism during this phase—with numerous tokens decoded per step—indicates the strong confidence of DLLM in generating the summarization content.

\textbf{Phase 2: Task Decomposition (Blocks 3-6).}
As the response proceeds, the planner agent conducts task decomposition during blocks 3-6. As shown in Figure~\ref{fig:planner_chart}(c), the number of tokens decoded per step decreases during this phase, indicating that the agent engages in more intensive reasoning and planning. Notably, despite the reduced parallelism and a decoding order that tends to proceed from left to right, DLLM still differs from AR models. While AR models generate content sequentially token by token, DLLM progressively fills in critical tokens to establish the reasoning framework before gradually completing the remaining parts of the response. Besides, Figure~\ref{fig:planner_chart}(b) shows a distinctive pattern: the entropy of remaining mask tokens may suddenly increase at the beginning or during intermediate steps within each block, reflecting moments of greater uncertainty in the planning process. As decoding progresses, the entropy generally decreases, signifying increased confidence in the generated content.

\noindent\textbf{Contrast with AR Planning Behavior.}
Unlike AR models that must commit to early tokens sequentially and cannot easily revise high-level decisions once formed, the DLLM's iterative refinement allows it to establish a global planning framework (parallel key point extraction) before filling in details. This contrasts with AR planners which typically extract points incrementally and may need to backtrack or reformulate plans if early commitments misalign with later analysis, leading to more intermediate planning artifacts (e.g., multiple \texttt{todo\_v1.md}, \texttt{todo\_v2.md} as observed in case studies in Section~\ref{sec:exp}) and redundant task decompositions.

To illustrate the qualitative differences, Table~\ref{tab:planner_comparison} presents a side-by-side comparison of key phrases from both DLLM and AR models' planning outputs. While both models extract the same four key query constraints, DLLM's task formulations are significantly more accurate and actionable. For example, whereas AR models use generic descriptions like \textit{"Chinese internet company parent"}, DLLM provides a more precise research scope: \textit{"Chinese internet companies that have created subsidiaries"}. Similarly, DLLM specifies \textit{"Mergers of in-house teams in tech companies"} instead of the vague \textit{"Formed by merging two teams"}. This higher specificity reduces ambiguity for downstream agent execution. Crucially, DLLM explicitly organizes tasks into a \textit{"structured plan"} with \textit{"parallel research tasks} to gather information from \textbf{multiple angles}" and directly assigns these to information seeker agents, demonstrating superior workflow coordination capabilities.

\begin{table}[ht]
\setlength{\abovecaptionskip}{2pt}
\setlength{\belowcaptionskip}{2pt}
\centering
\scriptsize
\begin{tabular}{@{}p{0.47\linewidth}@{}p{0.47\linewidth}@{}}
\toprule
\textbf{DLLM Planner Agent} & \textbf{AR Planner Agent} \\
\texttt{[User Intent Analysis] This is a complex query that requires research to identify the specific company. I need to \textbf{create a structured plan} to investigate this.} &
\texttt{[User Intent Analysis] This seems like a puzzle or riddle about identifying a specific tech company. The clues are quite specific, so I should research tech companies that match these criteria.} \\[1em]
\texttt{Let me break this down into \textbf{parallel research tasks} to gather information from \textbf{multiple angles}. \newline
\textbf{The key aspects to research are:} \newline
- Tech companies with animal-related names \newline
- Chinese internet companies that have created subsidiaries \newline
- Mergers of in-house teams in tech companies \newline
- Companies that do both software and hardware \newline
I should create a task planning document and then \textbf{assign parallel research tasks to information seeker agents}.} &
\texttt{Let me break this down: \newline
- Animal-named tech company \newline
- Chinese internet company parent \newline
- Formed by merging two teams \newline
- Software + hardware focus \newline
This requires research to find the exact company that matches all these criteria. I\'ll need to \textit{search for tech companies with animal names}, their parent companies, and the specific formation details.} \\
\bottomrule
\end{tabular}
\caption{Side-by-side comparison of planning outputs from dLLM and AR models. \texttt{[User Intent Analysis]} represents semantically similar content (4 key points) extracted by both models and is omitted for clarity. Key differences in planning approach are highlighted in bold.}
\label{tab:planner_comparison}
\end{table}


\noindent\textbf{Implications for Workflow Efficiency.}
The decoding patterns observed in the Planner Agent suggest that DLLM's efficiency advantage is not merely token-level parallelism. The ability to extract key query constraints in parallel (Phase 1) and establish reasoning frameworks before detailed completion (Phase 2) reflects a fundamentally different planning style: one that emphasizes global structural understanding before incremental elaboration. This reduces the need for multiple planning passes and intermediate artifact generation, explaining why DLLM planners produce more concise task assignments that information seekers can execute with fewer re-clarifications—consistent with the reduced information seeker agent invocation counts (8 vs. 10.4 calls per query) reported in Figure~\ref{fig:nseeker}.

\subsection{Information Seeker Agent}

\begin{figure}[htbp]
    \centering
    \includegraphics[width=0.85\linewidth]{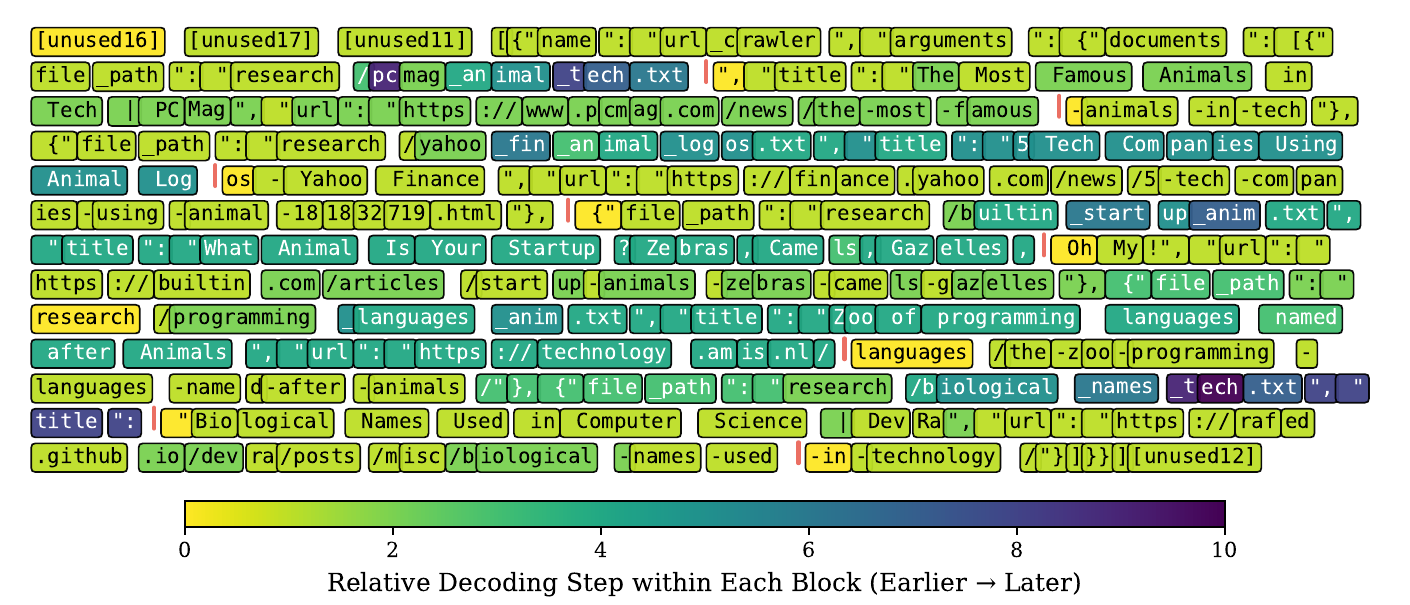}
    \caption{Heatmap showing the relative decoding order of each token within each block of an example Information Seeker Agent response (block size = 32). Different blocks are separated by red lines.}
    \label{fig:seeker_heatmap}
\end{figure}

The Information Seeker Agent executes the specific research tasks assigned by the planner, primarily focusing on tool invocation and information retrieval. As shown in the example response in Figure~\ref{fig:seeker_heatmap}, the seeker's behavior exhibits a distinctive \textbf{tool-oriented generation pattern} with high parallelism throughout the process.

\textbf{Tool Call Generation Pattern.}
The heatmap reveals a two-stage pattern in each tool invocation sequence. During the first block, the agent generates the decision to call the \code{url\_crawler} tool, establishing the overall action direction. Subsequent blocks then generate the specific parameters required for this tool call. This separation between tool selection and parameter specification reflects the agent's structured approach to information gathering, where high-level action planning precedes detailed argument construction.

\begin{figure}[htbp]
    \centering
    \includegraphics[width=1\linewidth]{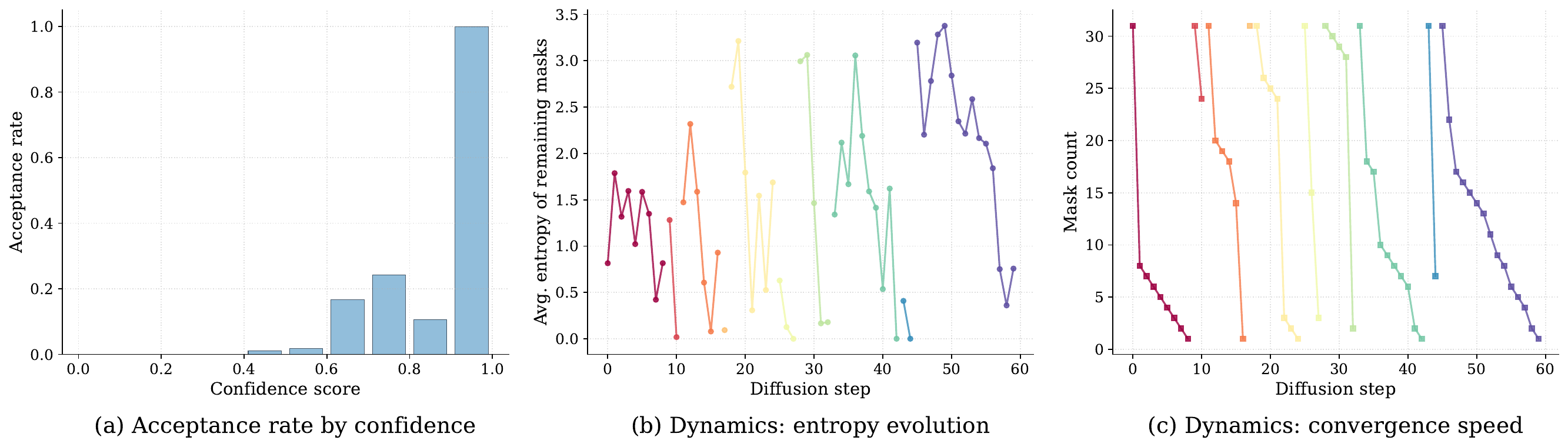}
    \caption{Analysis of the Information Seeker response shown in Figure~\ref{fig:seeker_heatmap}. (a) Probability of tokens with different confidence scores being decoded during diffusion process. (b) Average entropy of remaining mask tokens across different diffusion steps. (c) Number of remaining mask tokens within each block across different diffusion steps. In (b) and (c), different colors represent the decoding process of different blocks.}
    \label{fig:seeker_chart}
\end{figure}

\textbf{High Parallelism During Information Retrieval.}
Compared to the Planner Agent, the Information Seeker demonstrates significantly higher decoding parallelism. As information search tasks typically require less complex reasoning and decision-making, the agent can generate multiple tokens simultaneously within each diffusion step. Figure~\ref{fig:seeker_chart}(c) shows that the average number of tokens decoded per step remains consistently high throughout the entire response, confirming the seeker's efficient parallel processing capability.

\textbf{Uncertainty Dynamics Across Blocks.}
Figure~\ref{fig:seeker_chart}(b) reveals an interesting pattern of uncertainty within each block. At the beginning of each block, the model encounters increased uncertainty, indicated by higher entropy of remaining mask tokens. This reflects the initial exploration phase where the agent must decide which tool to invoke and how to structure the response. As decoding progresses within the block, uncertainty gradually decreases, signifying increased confidence as the agent fills in specific content and parameters. This within-block uncertainty pattern suggests that the most critical decisions---tool selection and action formulation---concentrate at the early stages of each generation block, followed by more deterministic content completion.

\noindent\textbf{Contrast with AR Tool-Call Generation.}
In AR-based agents, tool invocation typically proceeds linearly: function name and all parameters are generated sequentially without opportunity for revision or reorganization. If an early token decision (e.g., the function name or parameter structure) proves suboptimal, the agent must continue the sequence and may need to issue corrective calls in subsequent rounds, increasing interaction overhead. The DLLM's two-stage pattern—first establishing tool direction, then filling parameters through parallel refinement—allows the agent to globally coordinate the entire tool invocation before committing to execution. This reduces formulation errors that would require correction across interaction rounds, contributing to the cleaner tool-call traces observed in Section~\ref{sec:exp}'s case studies.

\subsection{Synthesis: Beyond Token-Level Parallelism}

Together, the behavioral patterns observed across both agent roles provide a coherent answer to our central question: the efficiency advantages of DLLM agents extend beyond token-level parallelism and stem from qualitatively different planning and tool-use behaviors.

Two paradigm-specific signatures emerge from our analysis:

\textbf{(1) Global Coordination Before Local Commitment.}
Both Planner and Information Seeker agents exhibit a pattern where critical high-level decisions are established early through parallel extraction or refinement, followed by deterministic completion of details. In the Planner Agent, this manifests as rapid parallel extraction of query constraints during the summarization phase, before task decomposition. In the Information Seeker, this appears as tool selection being finalized before parameter specification within each block. By contrast, AR agents must make irreversible local commitments from the beginning, requiring multiple refinement passes when early decisions misalign with emerging context. This explains why DLLM agents produce fewer intermediate artifacts and converge to valid trajectories with fewer interaction rounds.

\textbf{(2) Structured Action Formulation.}
The tool-oriented generation pattern in the Information Seeker reflects a diffusion-friendly interface where entire action spans are refined as structured units. This reduces formulation errors that would require correction across interaction rounds, whereas AR agents generate action components sequentially and must compensate through redundant verification loops—consistent with the heavier planning/verification patterns observed in AR case studies (Section~\ref{sec:exp}). The planner's entropy spikes during task decomposition further indicate that DLLM agents engage in more intensive global reasoning at the action-segment level, rather than distributing uncertainty across incremental token-level decisions.

\textbf{(3) Forward-aware Global Planning.}
Beyond local action formulation, the diffusion backbone exhibits a tendency to account for downstream and emerging workflow states when making each decision.
Through bidirectional conditioning over partially specified representations and repeated refinement, the model learns to jointly reconcile past context with anticipated future consequences.
This training dynamic potentially encourages decisions that are globally consistent across interaction rounds, rather than locally optimal under irreversible sequential commitments.

Together, these signatures indicate that the diffusion backbone's iterative refinement enables agents to ``plan'' at the \emph{action-segment level} rather than the \emph{token level}, systematically reducing workflow redundancy even after controlling for per-token generation speed. This paradigm shift—from sequential token commitments to structured segment refinement—provides the mechanistic explanation for the workflow-level efficiency gains observed in Section~\ref{sec:exp}.
We emphasize that these observations reflect tendencies exhibited under the current agent workflow, training setup, and evaluation regime.
While diffusion-based backbones demonstrate strong potential for forward-aware and globally consistent planning, their advantages may vary across tasks, model scales, and system configurations.
Exploring the generality of these behaviors in broader agent settings remains an important direction for future work.

%% file: arxiv/conclusion.tex
\section{\zhen{Open Problems and Future Directions}}
\label{sec:limitations_future}

While our empirical results show consistent efficiency gains of diffusion backbones within the DeepDiver-v2 workflow, several limitations and open research directions remain.

\paragraph{Scope and generalization.}
Our conclusions are drawn from controlled multi-turn, tool-augmented research-style benchmarks with fixed context windows and interaction budgets, which enables a clean backbone-level comparison.
It remains open whether the same efficiency and planning patterns persist in longer-horizon workflows (e.g., hundreds of turns), partial observability, richer and more heterogeneous tool ecosystems, or embodied environments where actions have delayed and irreversible consequences.
Establishing robustness across such regimes is essential to assess the practical impact of diffusion-based agents beyond benchmark settings.

\paragraph{Structured action reliability under strict parsers.}
Diffusion-based agents exhibit a higher invalid-action rate when tool calls must satisfy strict schemas.
Because diffusion generates action spans holistically through iterative refinement (rather than left-to-right conditioning), ensuring delimiter balance, argument typing, and schema compliance remains a deployment bottleneck.
Although many malformed spans can be corrected in later turns without reducing final accuracy, invalid actions still waste budget and may be unacceptable in safety-critical tool-use settings.
Future work should study structure-aware objectives (e.g., schema-constrained denoising, explicit bracket/field constraints) and decoding mechanisms (e.g., constrained sampling or grammar-guided refinement) tailored to diffusion backbones.

\paragraph{Exploration-efficiency trade-offs and long-horizon credit assignment.}
Our case studies suggest diffusion agents often follow more concise, contract-driven trajectories.
While this can improve efficiency, it may reduce exploratory breadth in tasks that require hypothesis branching, multi-path verification, or backtracking under uncertainty.
Characterizing when refinement-based policies under-explore, and how to introduce calibrated exploration (e.g., diversity in plan candidates or stochasticity schedules) without sacrificing efficiency, remains open.
A related challenge is long-horizon credit assignment: early planning choices can shape downstream tool usage and final correctness, yet diffusion distributes computation across denoising steps, making error attribution across turns and refinement stages less transparent than in token-factorized autoregressive models.

\paragraph{Reinforcement learning for diffusion-based tool agents.}
Scaling diffusion agents with RL introduces additional challenges beyond supervised trajectory learning.
First, interactive tool environments make on-policy data expensive, and naive RL may negate efficiency gains due to high sample complexity.
Second, structured tool actions impose hard validity constraints: penalties for invalid calls create sparse or discontinuous rewards, and small changes in an action span can flip a call from valid to invalid.
Third, diffusion introduces a two-timescale optimization problem---agent turns versus denoising steps---complicating advantage estimation and credit assignment within a single action.
Finally, offline or batch RL from logged tool trajectories must address distribution shift and confounding (e.g., the logging policy's tool choices), especially when evaluating counterfactual tool sequences.
Developing constraint-aware RL objectives, efficient off-policy training, and reliable evaluation protocols for diffusion-based tool agents is an important open direction.

\paragraph{Diffusion-native objectives and hybrid backbone designs.}
Our current training relies on continued learning from trajectories originally generated by autoregressive agents, which may limit diffusion-specific benefits.
Future work could explore diffusion-native supervision that explicitly leverages iterative refinement, intermediate planning states, trajectory-level noise schedules, or structured denoising losses aligned with tool schemas.
In addition, the complementary strengths of diffusion and autoregressive backbones motivate hybrid agents: diffusion for global planning or hypothesis consolidation, and autoregressive decoding for fine-grained execution and schema-critical tool interaction.
Designing such hybrids---including interfaces, arbitration, and shared memory---may yield robustness and efficiency gains beyond either paradigm alone.

\section{Conclusions}
\label{sec:conclusion}

\zhenew{In this work, we investigate how fundamental differences between diffusion LLMs and autoregressive (AR) LLMs manifest in tool-using agent behavior. 
Rather than proposing a new agent framework, we instantiate both backbones within the same DeepDiver workflow and perform matched continued training on identical tool-use trajectories. 
This controlled setup allows us to attribute behavioral differences to backbone-induced decision dynamics rather than system-level confounders.}

\zhenew{Our results show that diffusion-style modeling does not merely alter token generation mechanics; it changes how and when reasoning commitments are formed in multi-step agent workflows. 
Under identical interaction budgets and context constraints, the resulting \emph{DLLM Agent} achieves comparable final accuracy while improving end-to-end efficiency by over 30\% on average. 
Conditioned on success, diffusion-backed trajectories exhibit fewer conflict-correction events, reduced backtracking, and earlier global alignment of constraints, leading to shorter and more coherent tool-call traces. 
These gains arise at the workflow level rather than from parallel decoding alone.}

\zhenew{Beyond efficiency, we uncover a structural interaction between generation paradigm and task topology. 
Autoregressive decoding aligns naturally with linear multi-hop chains that favor sequential dependency propagation, whereas diffusion-style generation shows clear advantages on constraint-intersection and evidence-reconciliation problems that require simultaneous multi-constraint coordination and cross-source alignment. 
This suggests that generation paradigms induce distinct structural inductive biases in multi-turn reasoning, rather than constituting a simple speed–accuracy trade-off.
At the same time, diffusion backbones introduce practical challenges, including higher susceptibility to structured tool-call violations and sensitivity to context–action masking alignment. 
Addressing these issues through structure-aware training and decoding strategies is essential for reliable deployment.}

\zhenew{Overall, our findings indicate that diffusion backbones represent a viable and conceptually distinct alternative for tool-using agents. 
By shifting reasoning from incremental commitment toward earlier global coordination, diffusion-based agents offer a different allocation of decision-making effort across trajectories. 
We hope this work motivates further exploration of hybrid and structure-aware agent architectures that leverage complementary inductive biases across generation paradigms.}

%% file: arxiv/appendix.tex
\section{Additional Case Study}\label{subsec: case}

In this subsection, we present certain representative case studies where both backbones (AR and vanilla DLLM) produce the correct answer for the exact same browsing query, yet the end-to-end latency differs substantially. By inspecting the execution logs side by side, we highlight the behavioral differences in planning, tool-call patterns, and evidence consolidation that account for the speed advantage of the DLLM agent. 

\subsubsection{Multi-constraint Matching}

\begin{casebox}
\textbf{Case Study: DLLM agent on a multi-constraint browsing query.}

\medskip
\textbf{User Query.}
\begin{lstlisting}[style=logstyle]
A tech company: its name comes from an animal; its parent company is a Chinese
internet company; it was formed by merging two in-house teams; and it involves
both software and hardware. What is the company called?
\end{lstlisting}

\textbf{Planner $\rightarrow$ InformationSeeker task delegation (translated excerpt).}
\begin{lstlisting}[style=logstyle]
Action: assign_multi_objective_tasks_to_info_seeker -> Success
{
  "tasks": [{
    "task_content": "Search for a tech company that matches all constraints:
      (1) its name comes from an animal;
      (2) its parent company is a Chinese internet company;
      (3) it was formed by merging two in-house teams;
      (4) it involves both software and hardware.
      Provide the exact company name and relevant background.",
    "deliverable_contents": "Return the company that satisfies all constraints,
      including: (1) legal company name; (2) the animal-name origin;
      (3) parent-company information; (4) merger background (two teams);
      (5) business scope (software and hardware). Present results in a
      structured format.",
    "acceptance_checking_criteria": "You must find at least one company that
      fully satisfies all four constraints and provide sufficient evidence.
      If no perfect match is found, report the closest candidates and explain
      why they do not match.",
    "current_task_status": "Initial task: start from scratch to search for a
      company matching all constraints."
  }]
}
\end{lstlisting}

\textbf{Evidence reading + termination (translated excerpt).}
\begin{lstlisting}[style=logstyle]
Action: file_read -> Success {"file_path": "research/findings_summary.md"}

Action: planner_objective_task_done -> Success
{
  "completion_status": "completed",
  "final_answer": "Based on the provided constraints, the company is:
    T-Head Semiconductor Co., Ltd. ...",
  "key_files": [{"file_path": "research/findings_summary.md",
                 "is_final_output_file": true}]
}
\end{lstlisting}

\textbf{Tool-call trace (episode-level).}
\begin{lstlisting}[style=logstyle]
1) think
2) assign_multi_objective_tasks_to_info_seeker
3) file_read
4) planner_objective_task_done
\end{lstlisting} 

\textbf{Final answer.}
\begin{lstlisting}[style=logstyle]
final_answer:
Based on your described constraints, the company is T-Head Semiconductor
(Pingtouge).
\end{lstlisting}

\textbf{End-to-end latency on the same query (both correct): 140.95 s}

\end{casebox}

\begin{casebox}
\textbf{Case Study: AR agent on the same multi-constraint browsing query.}

\medskip
\textbf{User Query.}
\begin{lstlisting}[style=logstyle]
A tech company: its name comes from an animal; its parent company is a Chinese
internet company; it was formed by merging two in-house teams; and it involves
both software and hardware. What is the company called?
\end{lstlisting}

\textbf{Artifacts produced .}
\begin{lstlisting}[style=logstyle]
key_files:
- todo_v1.md (task plan v1; decomposition and execution plan)
- todo_v2.md (task plan v2; verification-focused plan)
- research/china_internet_companies_animal_subcompanies_report.md
  (report: 23 animal-named Alibaba subsidiaries)
- research/tech_company_team_mergers_report.md
  (report: 12 examples of tech companies formed by team mergers)
- research/candidate_companies_report.md
  (final report: confirms T-Head/Alibaba is the best match) [final output]
\end{lstlisting}

\textbf{Process characteristics.}
\begin{lstlisting}[style=logstyle]
Phase 1: Parallel multi-angle search (3 info-seeking agents dispatched)
  (1) animal-named subsidiaries of Chinese internet companies: succeeded
  (2) animal-named tech company search: failed (not finished within 30 iters)
  (3) companies formed by team mergers: succeeded

Phase 2: Focused verification (todo_v2.md)
  (1) detailed verification of T-Head: failed (not finished within 30 iters)
  (2) verification of other Alibaba subsidiaries: failed (<=30 iters)
  (3) alternative candidate search: succeeded

Phase 3: Integration and confirmation from completed reports; finalize answer.
\end{lstlisting}

\textbf{Tool usage summary.}
\begin{lstlisting}[style=logstyle]
- Planning documents: 2 versions (todo_v1.md, todo_v2.md)
- Parallel search: 6 tasks dispatched via assign_multi_objective_tasks_to_info_seeker
- File management: multiple reports generated
- Integration: file_read used to confirm final answer from reports
\end{lstlisting} 

\textbf{Final answer.}
\begin{lstlisting}[style=logstyle]
final_answer:
Based on your described constraints, the company is T-Head Semiconductor
(Pingtouge).

Details:
- Company: T-Head Semiconductor Co., Ltd. (T-Head Semiconductor)
- Parent company: Alibaba Group (Chinese internet company)
- Founded: 2018
- Name origin: "Pingtouge" is a nickname for the honey badger (an animal)
- Merger background: formed by merging two Alibaba teams
  (C-SKY Microsystems / Zhongtianwei, and DAMO Academy in-house chip team)
- Scope: chip R&D and design; software-hardware co-design / co-optimization
\end{lstlisting}

\textbf{End-to-end latency on the same query (both correct): 1152.68 s}
\end{casebox}

These two episodes are intentionally comparable: the AR agent and the DLLM agent solve the \emph{same} multi-constraint query and both return the correct entity (T-Head Semiconductor / Pingtouge). The difference lies in how each backbone behaves during multi-turn tool-based execution, which directly impacts end-to-end latency. Empirically, the vanilla DLLM agent finishes in 140.95s, while the AR agent requires 1152.68s on the identical query (an \textbf{8.18$\times$ wall-clock speedup} for DLLM). 

The log excerpts reveal complementary strengths and a clear efficiency trade-off. On the AR side, the PlannerAgent exhibits a thorough but heavy workflow: it creates multiple planning artifacts (\texttt{todo\_v1.md}, \texttt{todo\_v2.md}), dispatches 6 information-seeking tasks, and produces several intermediate research reports before integrating them into a final candidate report. Importantly, the AR summary also records repeated sub-task failures due to iteration caps (multiple tasks not completed within 30 iterations), after which the agent still succeeds by relying on other completed reports. This pattern is robust but computationally expensive: more planning rounds, more intermediate artifact generation, and longer-running verification loops accumulate substantial latency.

In contrast, the DLLM episode demonstrates a diffusion-friendly execution profile: it quickly compiles the query into an explicit multi-objective research contract (deliverables plus acceptance criteria) and closes the loop with a short, stable tool-call trace
(\texttt{think} $\rightarrow$ \texttt{assign} $\rightarrow$ \texttt{file\_read} $\rightarrow$ \texttt{done}).
This matters because diffusion-style action generation can operate with a bounded denoising budget per turn and does not require long autoregressive decoding to emit structured tool calls. In multi-turn browsing---where each round must output a well-formed, parseable action span---this fixed-compute, refinement-based generation can translate into lower per-turn overhead and fewer opportunities for protracted verification loops, yielding a large reduction in wall-clock time while preserving correctness. Overall, the side-by-side logs suggest that the diffusion backbone’s efficiency advantage manifests not only in faster generation, but also in a more direct, contract-driven execution path that reduces interaction and coordination overhead.

This second case study complements the earlier BrowseComp-zh episode by switching from a constraint-satisfaction browsing task to an \emph{open-ended report-generation} task (``analyze Tesla 2024 Q3 earnings; estimate China deliveries and market share; issue an investment rating''). Here, both agents again complete the task successfully, but the efficiency gap narrows in a predictable way. The DLLM agent finishes end to end in 490.25s, whereas the AR agent requires 715.31s (1.46$\times$ latency; equivalently, DLLM is 31.5\% faster).

\subsubsection{Open Questions}

\begin{casebox}
\textbf{Case Study: DLLM agent on the report-generation query.}

\medskip
\textbf{User Query.}
\begin{lstlisting}[style=logstyle]
Analyze Tesla's 2024 Q3 earnings report. Look up China market deliveries and
market share, and then provide an investment rating.
\end{lstlisting}

\textbf{Planner -> InformationSeeker delegation (contract-style; excerpt).}
\begin{lstlisting}[style=logstyle]
Action: assign_multi_objective_tasks_to_info_seeker -> Success
{
  "tasks": [{
    "task_content": "Obtain Tesla 2024 Q3 earnings materials (shareholder deck,
      10-Q, earnings call transcript). Extract key financials and guidance.
      Cite sources and page/section anchors."
  },{
    "task_content": "Find Tesla China deliveries in 2024 Q3 (or closest proxy if
      Tesla does not disclose directly). Provide methodology and sources."
  },{
    "task_content": "Estimate China EV market share for Tesla in 2024 Q3 using
      a credible denominator (e.g., BEV passenger car sales). Provide the
      data source, definitions, and calculation."
  },{
    "task_content": "Collect consensus / street views and major risks.
      Provide 3-5 credible sources and summarize divergences."
  }],
  "deliverable_contents": "Return a publishable report with: executive summary,
    financial highlights, China deliveries + market-share calculation, thesis,
    valuation notes (if feasible), risks, and a final rating (Buy/Hold/Sell)
    with justification. Include a sources section.",
  "acceptance_checking_criteria": "Every major numeric claim must be sourced.
    If an item cannot be verified from credible sources, mark it as
    'not disclosed' and use a proxy with clear caveats."
}
\end{lstlisting}

\textbf{Evidence consolidation + report writing (excerpt).}
\begin{lstlisting}[style=logstyle]
Action: file_read -> Success {"file_path": "research/tesla_q3_2024_financials.md"}
Action: file_read -> Success {"file_path": "research/tesla_china_deliveries_proxy.md"}
Action: file_read -> Success {"file_path": "research/china_ev_market_share_calc.md"}
Action: file_read -> Success {"file_path": "research/risks_and_consensus.md"}

Action: planner_objective_task_done -> Success
{
  "completion_status": "completed",
  "final_answer": "Investment Memo: Tesla 2024 Q3 (Report)\n\nExecutive Summary:\n- ...\n\nChina deliveries & share:\n- ... (with sources and caveats)\n\nRating: HOLD\nRationale:\n- ...",
  "key_files": [
    {"file_path": "reports/tsla_2024q3_investment_memo.md", "is_final_output_file": true}
  ]
}
\end{lstlisting}

\textbf{Tool-call trace (episode-level).}
\begin{lstlisting}[style=logstyle]
1) think
2) assign_multi_objective_tasks_to_info_seeker
3) file_read (x4)
4) planner_objective_task_done
\end{lstlisting}

\textbf{Note (why the speedup is smaller here).}
\begin{lstlisting}[style=logstyle]
Open-ended report generation requires substantially more long-form writing.
Even if per-token generation is faster, end-to-end advantage can shrink when
total output length and evidence narration dominate runtime.
\end{lstlisting} 

\textbf{End-to-end latency on the same query (both correct): 490.25 s}
\end{casebox}

\begin{casebox}
\textbf{Case Study: AR agent on the report-generation query.}

\medskip
\textbf{User Query.}
\begin{lstlisting}[style=logstyle]
Analyze Tesla's 2024 Q3 earnings report. Look up China market deliveries and
market share, and then provide an investment rating.
\end{lstlisting}

\textbf{Planning + verification-heavy workflow (excerpt).}
\begin{lstlisting}[style=logstyle]
Action: file_write -> Success {"file_path":"todo_v1.md", "desc":"plan v1"}
Action: assign_multi_objective_tasks_to_info_seeker -> Success {"tasks":[...]}
Action: file_write -> Success {"file_path":"todo_v2.md", "desc":"plan v2 (verification)"}

Action: file_read -> Success {"file_path":"research/tesla_q3_2024_financials.md"}
Action: file_read -> Success {"file_path":"research/tesla_china_deliveries_proxy.md"}
Action: file_read -> Success {"file_path":"research/china_ev_market_share_calc.md"}

Action: assign_multi_objective_tasks_to_info_seeker -> Success
{"tasks":[{
  "task_content":"Re-check China-deliveries proxy against an alternative source
    and reconcile discrepancies; if inconsistent, update the estimate and
    caveats."
},{
  "task_content":"Re-check market-share denominator definitions (BEV vs NEV vs
    passenger cars). Provide sensitivity analysis if needed."
}]}
\end{lstlisting}

\textbf{Termination with long-form report (excerpt).}
\begin{lstlisting}[style=logstyle]
Action: planner_objective_task_done -> Success
{
  "completion_status": "completed",
  "final_answer": "Investment Memo: Tesla 2024 Q3 (Report)\n\nExecutive Summary:\n- ...\n\nChina deliveries & share:\n- ... (sources, caveats, sensitivity)\n\nRating: HOLD\nRationale:\n- ...",
  "key_files": [
    {"file_path": "todo_v1.md", "is_final_output_file": false},
    {"file_path": "todo_v2.md", "is_final_output_file": false},
    {"file_path": "reports/tsla_2024q3_investment_memo.md", "is_final_output_file": true}
  ]
}
\end{lstlisting}

\textbf{Tool-call trace (episode-level).}
\begin{lstlisting}[style=logstyle]
1) think
2) file_write (todo_v1)
3) assign_multi_objective_tasks_to_info_seeker
4) file_write (todo_v2)
5) file_read (x3)
6) assign_multi_objective_tasks_to_info_seeker (re-check)
7) planner_objective_task_done
\end{lstlisting}

\textbf{End-to-end latency on the same query (both correct):  715.31 s}
\end{casebox}

The key contrast with the first case is that report generation is dominated less by discrete tool-action decisions and more by \emph{long-form synthesis}. Even if diffusion-style action generation reduces overhead for producing well-formed tool calls and high-level plans, the runtime share attributable to writing (multi-section narratives, caveats, methodology explanations, and structured ``memo'' formatting) increases substantially. As a result, the per-turn/per-action efficiency advantage of the diffusion backbone is partially amortized by the unavoidable cost of generating a much larger volume of final text and weaving together multiple evidence streams.

Nevertheless, the DLLM agent retains a clear end-to-end advantage. Consistent with the execution patterns observed in the first case, the diffusion-backed agent tends to converge earlier to a stable ``research contract'' (deliverables plus acceptance criteria) and proceeds with a more direct retrieve-then-write trajectory, whereas the AR agent more frequently expands the workflow with additional planning and verification passes (e.g., re-checking definitions/denominators for market share and reconciling alternative sources). In short, when tasks are short and path-finding dominates, diffusion yields dramatic speedups (as in BrowseComp-zh); when tasks are open-ended and output-length dominates, the advantage persists but naturally compresses (490.25s vs.\ 715.31s), indicating that \textbf{DLLM efficiency gains arise at the \emph{agent-workflow level} rather than solely from parallel token emission}.